\documentclass{article}

\usepackage{amsmath,amsfonts,bm}

\def\eqref#1{equation~\ref{#1}}

\def\1{\bm{1}}

\DeclareMathAlphabet{\mathsfit}{\encodingdefault}{\sfdefault}{m}{sl}
\SetMathAlphabet{\mathsfit}{bold}{\encodingdefault}{\sfdefault}{bx}{n}

\def\gA{{\mathcal{A}}}

\def\gM{{\mathcal{M}}}

\def\gP{{\mathcal{P}}}

\def\gS{{\mathcal{S}}}

\usepackage{arxiv}
\usepackage{url}
\usepackage{fancyhdr}
\usepackage{graphicx}

\usepackage{multirow}
\usepackage{diagbox}
\usepackage{algorithm}
\usepackage[noend]{algpseudocode}
\usepackage{subfigure}
\usepackage[T1]{fontenc}

\RequirePackage{algorithm}
\usepackage[backref]{hyperref}

\usepackage{hyperref}
\usepackage{url}
\usepackage{graphicx}
\usepackage{booktabs}
\usepackage{multirow}
\usepackage{makecell}
\usepackage{mdframed}
\usepackage{wrapfig}
\usepackage{xcolor}

\newcommand{\commenttext}[1]{\textcolor{light-gray}{#1}}
\newcommand{\llmcompletion}[1]{\colorbox{highlight}{#1}}

\definecolor{dark-gray}{HTML}{A9A9A9} %
\definecolor{light-gray}{HTML}{b7b7b7} %
\definecolor{light-green}{HTML}{dcf8c6}
\definecolor{light-blue}{HTML}{a5cdfb}
\definecolor{bg-gray}{HTML}{F8F8F8} 
	
\definecolor{dark-green}{HTML}{6aa84f} 
\definecolor{highlight}{HTML}{cfe2f3} 
\definecolor{blue}{HTML}{2B60DE} 
\definecolor{brown}{HTML}{9F8C76}
\definecolor{pink}{HTML}{D88782}
\definecolor{dark-yellow}{HTML}{F6AE2D}
\definecolor{light-yellow}{HTML}{ebdc7d}


\definecolor{lightgray}{gray}{0.893}
\usepackage{colortbl}

\usepackage{mathtools} 
\usepackage{booktabs} 
\usepackage{tikz} 
\usepackage{bm}
\title{Efficient Human-AI Coordination via\\Preparatory Language-based Convention}

\author{%
 Cong Guan$^{1}$, Lichao Zhang$^2$, Chunpeng Fan$^2$, Yichen Li$^1$, Feng Chen$^1$, Lihe Li$^1$,Yunjia Tian$^2$,\\ \textbf{Lei Yuan$^{1,2}$, {Yang Yu}$^{1,2}$}\thanks{Corresponding Author}\\
  $^1$ National Key Laboratory for Novel Software Technology, Nanjing University\\
  $^2$ Polixir.ai\\
  \texttt{guanc@lamda.nju.edu.cn}, \texttt{\{lichao.zhang,chunpeng.fan\}@polixir.ai}, \\\texttt{\{liyc,chenf,lilh\}@lamda.nju.edu.cn}, \texttt{yunjia.tian@polixir.ai},\\ \texttt{yuanl@lamda.nju.edu.cn}, \texttt{yuy@nju.edu.cn}
}

\begin{document}
\date{}
\maketitle
\begin{abstract}
Developing intelligent agents capable of seamless coordination with humans is a critical step towards achieving artificial general intelligence. Existing methods for human-AI coordination typically train an agent to coordinate with a diverse set of policies or with human models fitted from real human data. However, the massively diverse styles of human behavior present obstacles for AI systems with constrained capacity, while high quality human data may not be readily available in real-world scenarios. In this study, we observe that prior to coordination, humans engage in communication to establish \textit{conventions} that specify individual roles and actions, making their coordination proceed in an orderly manner. Building upon this observation, we propose employing the large language model (LLM) to develop an action plan (or equivalently, a convention) that effectively guides both human and AI. By inputting task requirements, human preferences, the number of agents, and other pertinent information into the LLM, it can generate a comprehensive convention that facilitates a clear understanding of tasks and responsibilities for all parties involved. Furthermore, we demonstrate that decomposing the convention formulation problem into sub-problems with \textit{multiple} new sessions being sequentially employed and human feedback, will yield a more efficient coordination convention. Experimental evaluations conducted in the \textit{Overcooked-AI} environment, utilizing a human proxy model, highlight the superior performance of our proposed method compared to existing learning-based approaches. When coordinating with real humans, our method achieves better alignment with human preferences and an average performance improvement of 15\% compared to the state-of-the-art.
\end{abstract}

\section{Introduction}
Training intelligent agents that can effectively coordinate with humans~\cite{overcooked-ai} is crucial for enhancing productivity in human society and represents one of the most significant challenges in the pursuit of artificial general intelligence~\cite{goertzel2007artificial,DBLP:journals/chb/Endsley23}. Previous approaches to human-AI coordination can be broadly classified into three main directions~\cite{instructRL}. The first direction involves directly fitting human behaviors or intentions using real human data~\cite{DBLP:journals/corr/abs-2210-05125, parekh2023learning}. The second direction focuses on designing algorithms or reward functions inspired by cognitive science to generate human-like policies~\cite{DBLP:conf/icml/HuLCPBF21, DBLP:conf/nips/CuiHPF21, DBLP:conf/iclr/LaidlawD22, hsp}. The third direction, known as Population-Based Training (PBT)~\cite{jaderberg2017population}, entails constructing a diverse pool of teammates and training a common best response policy~\cite{fcp, mep}. In recent years, researchers have developed various algorithms around these directions, leading to remarkable advancements in human-AI coordination across various domains, including industrial assembly lines~\cite{DBLP:journals/cor/NourmohammadiFN22}, healthcare~\cite{DBLP:conf/hci/GleichaufSW22}, and video games~\cite{DBLP:conf/nips/SiuPCZLPCA21}, etc.

However, existing methods in these directions have certain limitations. In real-world scenarios, obtaining high-quality human data is not always easily accessible, which hampers the feasibility of fitting human models~\cite{strouse2021collaborating}. Moreover, the mechanisms underlying human behaviors are complex~\cite{thatcher2021foundations}. Although incorporating limited representative factors such as attention and irrationality into algorithm design can produce policies that resemble humans more than vanilla reinforcement learning~\cite{sutton2018reinforcement}, it still falls considerably short of capturing true human policies~\cite{DBLP:conf/iclr/LaidlawD22}. Benefiting from the widely proven instance generation ability of PBT~\cite{jaderberg2017population}, many attempts been successfully applied~\cite{strouse2021collaborating,mep,hsp,xue2022heterogeneous,charakorn2022generating}. 
Despite of this, the challenges associated with PBT methods include: Firstly, maintaining a diverse pool of teammates is not a trivial task, as the policies within the pool need to exhibit sufficient diversity and cover a wide range of real human policies~\cite{hsp}. Secondly, the policies trained through PBT only have experience in coordinating with teammates from the pool, resulting in poor generalization when encountering unseen teammates~\cite{fastap}. Note that we humans, before coordination, often reach a \textit{convention}~\cite{shih2020critical,gordon2023impact}\footnote{We use ``convention" to refer to the action plan for humans and AIs.} through communication to determine each individual's task and how to coordinate with each other. Therefore, a natural question arises: \textit{Can we enhance human-AI coordination via human-AI communication?} This is not trivial since humans excel at communicating using natural language, whereas AIs are not proficient in it~\cite{weng2020odqa}. None of existing methods of the above three directions have the potential to deal with this issue, so we need to find a new way for humans and AIs to discuss and coordinate.

\begin{figure}[t]
    \centering
    \includegraphics[width=0.87\linewidth]{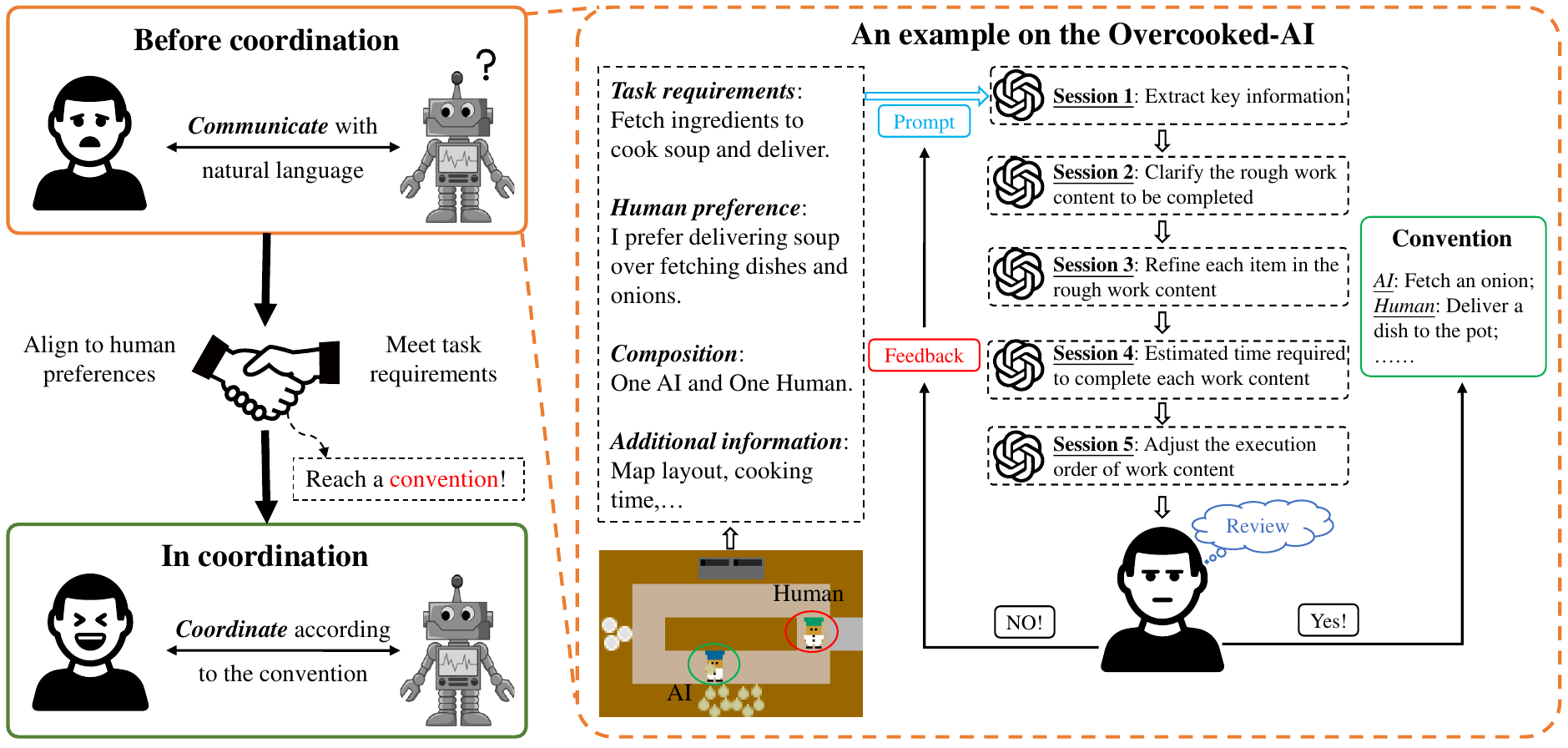}
    \caption{Overview of our proposed HAPLAN on the Overcooked-AI.}
    \label{fig:overview}
\end{figure}

Recently, Large Language Models (LLMs)~\cite{survey} have demonstrated impressive capabilities of natural language processing and task planning across various domains, such as robot control~\cite{DBLP:journals/corr/abs-2303-00855}, reasoning~\cite{DBLP:conf/acl/QiaoO0CYDTHC23}, and long conversation understanding~\cite{DBLP:conf/acl/LeeHPP023}, indicating their potential as bridges for human-AI communication and coordination. Hence, this work considers whether we can effectively apply LLMs to facilitate human-AI coordination. A naive approach is to input task requirements, human preferences, the number of AI agents, and other relevant information into an LLM before human-AI coordination, and request it to devise a convention, based on which humans and AIs will coordinate. We conducted experiments on this idea (please see \ref{sec:exp} for more details), but found it not work well, especially for hard scenarios. We suspect that although the reasoning and planning abilities of the current LLMs are greatly enhanced by emergent techniques such as Chain-of-Thought~\cite{chain-of-thought}, Least-to-Most~\cite{least-to-most}, it still suffers when dealing with challenging problems. When faced with complex tasks or lengthy conversation histories, it struggles to handle them well, and will generate inefficient conventions due to model hallucination~\cite{hallucination}.

To tackle the above issues, we propose efficient \textbf{H}uman-\textbf{A}I coordination via \textbf{P}reparatory \textbf{LA}nguage-based conventio\textbf{N} (\textbf{HAPLAN}), a novel framework for human-AI coordination via language-based conventions. When meeting a new task, HAPLAN will first decompose the formulation of a convention into several sub-problems and allocate each of them to a new session separately. Sequentially, the solution of one sub-problem will be the input of another session, which as a result develops a convention specifying roles and assignments for all involved parties. Additionally, to mitigate mistakes made by the LLM, HAPLAN will ask humans to inspect the proposed convention. If any issues are identified, humans will provide feedback to the LLM and require it to reformulate a convention, realizing explicit human-ai bidirectional value alignment~\cite{yuan2022situ}. \ref{fig:overview} illustrates an example of our proposed HAPLAN. Comparing with the above naive approach, HAPLAN has at least two advantages: First, each session only needs to handle a simple sub-problem, reducing the probability of making mistakes. Second, by leveraging human feedback, we can refine the planning results from the LLM, thereby improving the effectiveness of human-AI coordination. To evaluate the effectiveness of our approach, we conduct extensive experiments with human proxy models on five maps from Overcooked-AI~\cite{overcooked-ai}, a generally used benchmark for human-AI coordination. The results show that HAPLAN significantly outperforms existing approaches. Furthermore, by inviting real human players, we observe that HAPLAN achieves higher performance and a better alignment with different humans' preferences. Surprisingly, we find that the idea of decomposing a problem into several sub-problems and assigning them to different sessions can benefit other domains besides human-AI coordination. Experiments on benchmarks of symbolic manipulation, compositional reasoning and math reasoning demonstrate the generality of our idea.

\section{Related Work}

\paragraph{Human-AI Coordination}
Existing works on human-AI coordination can be broadly categorized into three main directions~\cite{instructRL}. The first direction is to model human behaviors and biases from real human data~\cite{overcooked-ai, DBLP:journals/corr/abs-2210-05125}. However, high-quality human data may not be readily available before human-AI coordination in real-world scenarios. In this work, we consider the setting where there are \textit{no} data of human-AI, human-human, AI-AI coordination. The second direction focuses on designing algorithms or reward functions inspired by cognitive science to generate human-like policies~\cite{DBLP:conf/icml/HuLCPBF21, DBLP:conf/nips/CuiHPF21, DBLP:conf/iclr/LaidlawD22, hsp}. Nevertheless, human behaviors are determined by various factors and complex mechanisms~\cite{thatcher2021foundations}. Although taking things like irrationality~\cite{DBLP:conf/iclr/LaidlawD22}, risk sensitivity~\cite{rmix} into consideration will generate policies that resemble humans more than vanilla RL, it is still difficult to fully capture the characteristics of human behaviors. Different from them, our method will ask humans to give their preferences to the LLM and review the proposed conventions, ensuring an effective human-AI coordination with a better alignment to human biases. The third direction, known as Population-Based Training (PBT), entails constructing a diverse pool of teammates and training a common best response policy~\cite{fcp, mep}. Maintaining the diversity of teammates pool under the requirement of covering human policies is not trivial. Moreover, there is no guarantee on the generalizability of the trained policy to unseen humans since it has only coordinated with teammates from the pool. 

\paragraph{Reasoning and Task Planning via LLMs}
Recently, Large Language Models (LLMs) has emerged as powerful tools in different domains~\cite{survey}. Reasoning is an essential ability for complex problem-solving~\cite{DBLP:conf/acl/QiaoO0CYDTHC23}. To improve the reasoning ability of LLMs, \cite{chain-of-thought} proposes Chain-of-Thought to encourage LLMs to explain their reasoning process. We have also taken this idea when designing prompts for multiple sessions   (please see \ref{sec:prompt} for more details). However, when solving problems harder than the exemplars shown in the prompts, Chain-of-Thought tends to perform poorly. \cite{least-to-most} proposes to break down a complex problem into a series of simpler sub-problems and then solve them in sequence, named as Least-to-Most. It requires the LLM to solve all the sub-problems in one session, while our method assigns each sub-problem to a new session separately. There are also some works considering utilizing LLMs to do task planning as our work does, such as \cite{DBLP:journals/corr/abs-2211-09935} and \cite{DBLP:conf/icml/HuangAPM22}. But none of them considers human-AI coordination tasks. Some works also try to enable multi-agent coordination with LLMs. \cite{sama} proposes to use an LLM to generate and assign sub-goals for AI-AI coordination. \cite{proagent} considers human-AI coordination, and proposes to integrate an LLM into the field of AI, serving to anticipate humans’ forthcoming decisions. We instead use an LLM to make conventions. For users who are interested in LLMs, we recommend to refer to up-to-update surveys such as \cite{survey}, \cite{llmagent} and \cite{surveyfd}.

\section{Preliminaries}

\paragraph{Two-Player Human-AI Cooperative Game}
In this work, we focus on two-player human-AI coordination, which can be modeled as a two-player Markov decision process extend form markov games~\cite{littman1994markov}, denoted by $\gM = \langle I, \mu_0, \gS, \boldsymbol{\gA}, \gP, R, \gamma \rangle$. Here, $I=\{A, H\}$ is the set of players, where we use $A$ to denote the AI and $H$ to denote the human. $\mu_0$ is the initial state distribution; $\gS$ is the state space; $\boldsymbol{\gA} = \gA^{  (A)}\times\gA^{  (H)}$ is the action space; $\gP: \gS \times \boldsymbol{\gA} \to \Delta_\gS$ is the transition function\footnote{We use $\Delta_X$ to denote the set of probabilities over $X$.}; $R: \gS\times\boldsymbol{\gA}\to\mathbb{R}$ is a global reward function shared by the human and the AI; $\gamma\in [0,1)$ is the discount factor. Let $\pi_A: \gS \to \Delta_{\gA^{  (A)}}$ be the AI's policy and $\pi_H: \gS \to \Delta_{\gA^{  (H)}}$ be the human's policy. We can define the expected discounted return as $J  (\pi_A, \pi_H) = \mathbb{E}\left[\sum_{t=0}^\infty \gamma^t R  (s_t, a_t^{  (A)}, a_t^{  (H)})\right]$, where $s_0 \sim \mu_0, a_t^{  (A)} \sim \pi_A  (\cdot|s_t), a_t^{  (H)} \sim \pi_H  (\cdot|s_t), s_{t+1} \sim \gP  (\cdot|s_t, \{a_t^{  (A)}, a_t^{  (H)}\})$. The goal is to specify $\pi_A$ and $\pi_H$ to achieve the highest $J  (\pi_A, \pi_H)$. Here, by saying ``specify'', we mean to develop a convention with human's preferences being satisfied for both human and AI. 

\paragraph{Convention-based Human-AI Coordination}
Inspired by human-human coordination, we consider making conventions for human-AI coordination. Specifically, a convention is a detailed action plan that assigns roles, tasks and other coordination details for both human and AI. Since LLMs are proficient in manipulating natural language, which is convenient for describing high-level plans, but not low-level control instructions, we use an LLM for human-AI coordination in a \textit{hierarchical} manner. That is, the convention proposed by the LLM describes high-level plans. To translate the convention into actions, human can harness his/her ability of natural language understanding, whereas AI relies on pre-trained low-level policies.

\section{Method}

This section describes details of our proposed method, HAPLAN. We will begin with an introduction to the prompt designation, the manipulation of multiple sessions and the whole pipeline. Then, we proceed to explain how to train the low-level policies.

\subsection{Task Planning with Multiple Sessions}

\begin{figure}[ht]
    \centering
    \includegraphics[width=0.9\linewidth]{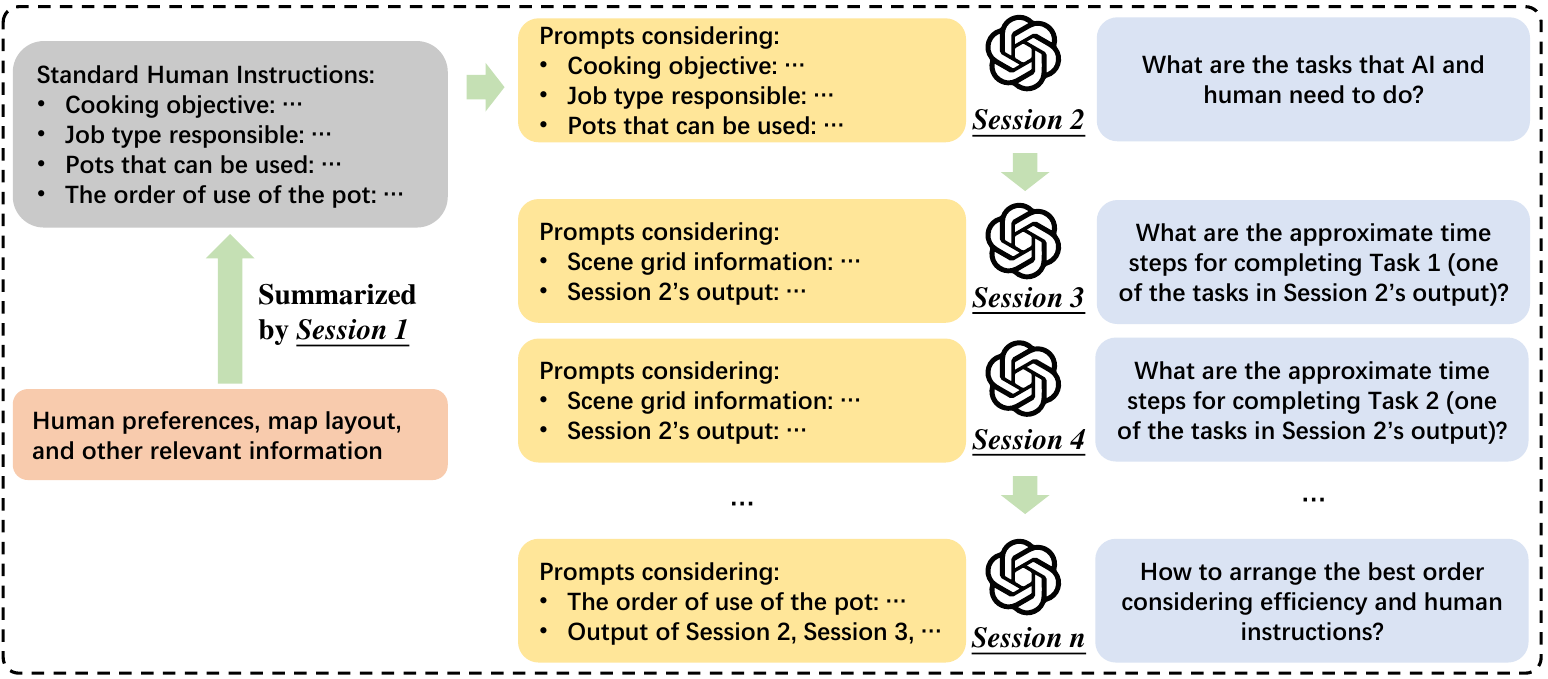}
    \caption{An example of task planning via multiple sessions.}
    \label{fig:planner}
\end{figure}

\begin{figure}[ht]
    \centering
    \includegraphics[width=0.95\linewidth]{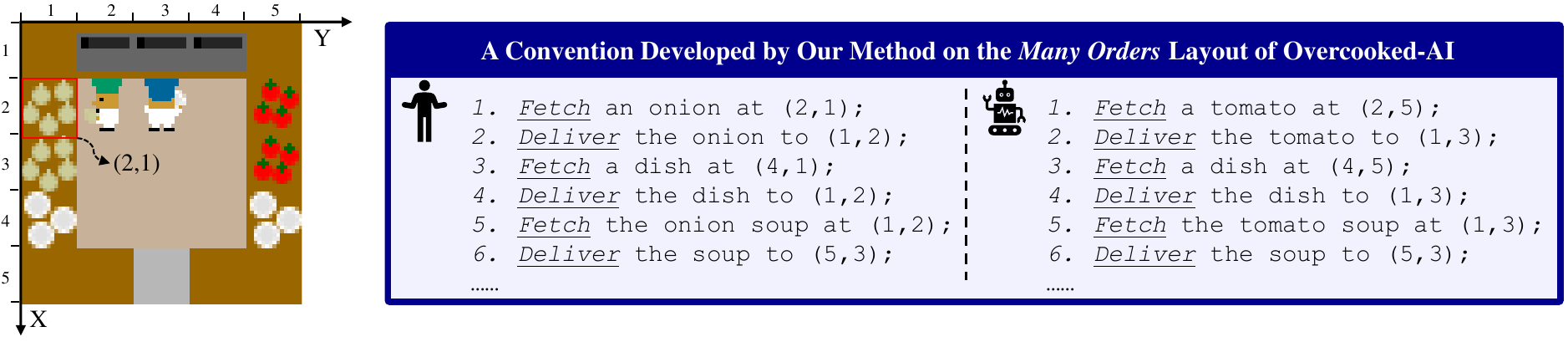}
    \caption{An example of conventions on Overcooked-AI. \textbf{Left}: Layout of the \textit{Many Orders} map; \textbf{Right}: A convention for human and AI, where the left part is action plans for human and the right part is action plans for AI. $(x,y)$ in the plans denotes the  region in the layout whose coordinate on the X-axis is $x$ and coordinate on the Y-axis is $y$.}
    \label{fig:convention}
\end{figure}

\paragraph{Planner based on Multiple Sessions}
Extensive empirical results have shown that when dealing with a complex problem or a long conversation history, the LLM may struggle to effectively cope, leading to the generation of misleading contents~\cite{hallucination}. Although methods such as Chain-of-Thought~\cite{chain-of-thought} and Least-to-Most~\cite{least-to-most} have greatly improved the reasoning capabilities of the LLM, our experiments in \ref{sec:exp} have revealed that it still faces challenges in addressing more difficult human-AI coordination tasks. To tackle this issue, we propose employing multiple new \textit{sessions} to jointly develop conventions. Specifically, we decompose a complex problem into multiple sub-problems and assign them sequentially to a new session. By doing so, in each session, the LLM only needs to a much simpler sub-problem and shorter prompt, thus alleviating the issue of model hallucination~\cite{hallucination}. The solution provided by one session serves as part of the prompt for the subsequent session. Similar to role play~\cite{roleplay}, we implement reasoning via multiple sessions by starting new sessions on ChatGPT~\cite{chatgpt}, each with a different prompt. A typical decomposition is shown in \ref{fig:planner}. \ref{fig:convention} illustrates a convention developed by the multiple sessions on the Overcooked-AI environment.

\paragraph{Re-plan from Human Feedback}
To ensure that the generated convention is both efficient and aligned with human preferences, we incorporate a human validation process. That is, human will review the generated content and provide feedback on any inappropriate aspects. Modifications suggested by human will then be used as part of the prompts for the \textit{first} session to re-plan the convention. We also provide an example of re-planning due to human feedback in \ref{sec:prompt}.

\subsection{Execution with Pre-Trained Skills}
The convention contains temporally extended high-level instructions in natural language, which has to be translated into low-level actions executable for AI. To do so, we have pre-trained several skills, similar to SayCan~\cite{saycan}. Taking the Overcooked-AI~\cite{overcooked-ai} environment as an example, we use the following two skills, visualed in \ref{sec:impl}:
\begin{itemize}
    \item \texttt{Fetch} This skill empowers the AI to take something at some place. Generally, when there is a need to fetch something, denoted as \texttt{A}, at some place, denoted as \texttt{B}, we require the LLMs to output a sentence like ``\texttt{Fetch A at B}''.
    \item \texttt{Deliver} This skill enable the AI to take something to some place. When there is a requirement to take something, denoted as \texttt{A}, to some place, denoted as \texttt{B}, the LLMs will output a plan like ``\texttt{Deliver A to B}''.
\end{itemize}

We train these two skills using behavioral cloning~\cite{bc} from human demonstrations. To increase the generalizabilities of the learned skills, we have AI adopt a random policy when collecting human demonstrations. For more implementation details, please refer to \ref{sec:impl}. Depending on the environments, we can also learn more different skills, so our approach is scalable.

\section{Experiments}\label{sec:exp}

To validate whether our proposed approach HAPLAN can indeed leverage the advantages of Large Language Models   (LLMs) to enhance Human-AI coordination to a new level, we choose Overcooked-AI~\cite{overcooked-ai}, a test environment commonly used in previous research on human-AI coordination, for empirical experiments. In this environment, there exist two players required to coordinate to complete several order tasks, with each order including a series of steps: fetching ingredients and placing them in the pot, cooking the soup, taking the dishes to scoop the soup and delivering the soup. 
Specifically, we select five layouts from the environment, and conduct test experiments with both human proxy models and real humans. More details about the Overcooked-AI environment and the selected layouts can be found in~\ref{appx:overcooked_ai_env}.
For baselines, we compare our approach with several popular human-AI coordination algorithms, respectively: Fictitious Co-Play (FCP)~\cite{fcp}, Maximum Entropy Population-based training (MEP)~\cite{mep} and Hidden-utility Self-Play (HSP)~\cite{hsp}. In~\ref{appx:baselines}, we introduce the details of these methods.

With the experiments,
we aim to answer the following questions: 1) Can our approach obtain better human-AI coordination performance than the existing traditional methods no matter when faced with human proxy models or real humans?   (See Section~\ref{exp:main_exp}) 2) Why can the inclusion of LLMs enhance the human-AI coordination performance and what does it bring about?   (See Section~\ref{exp:analysis}) 3) Does utilizing multiple sessions enhance the reasoning capability of LLMs?   (See~\ref{exp:ablation}) 4)   How does our approach perform on other reasoning benchmarks?   (See~\ref{exp:reasoning})

\subsection{LLMs Enhance Coordination Performance}
\label{exp:main_exp}

In this section, we conduct experiments to validate whether the inclusion of LLMs indeed benefits the human-AI coordination. We first introduce the experiments of coordination with human proxy models, and later provide details and results related to experiments with real humans. 

\subsubsection{Coordinating with Human Proxy Models}

Firstly, we want to test the ability of our approach to coordinate with partners of different coordination patterns. To serve this purpose, we adopt the scripted policies in HSP~\cite{hsp} as the testing partners, which have strong preferences in coordination patterns. To achieve good coordination with these scripted policies, the AI agent must recognize the partner's preference and adapt to it effectively. This poses great challenges to traditional methods, as they do not have an explicit process for knowing about the partner. The experimental results are presented in Table~\ref{tab:exp1}.

\begin{table}[t!]
\centering
\caption{Experimental results on Overcooked-AI environment 
of HAPLAN and baselines when coordinating with human proxy policies. The best values have been \textbf{bolded}.}
\label{tab:exp1}
\centering
\resizebox{\linewidth}{!}{
\begin{tabular}{c|c|c|c|c|c}
\toprule
    Layout  &       Partner     &               FCP &               MEP &               HSP &           HAPLAN \\
\midrule
\multirow{2}{*}{Counter Circle} & Onion Placement &   104.38$\pm$9.66 &  133.75$\pm$20.27 &  135.38$\pm$15.19 &  \textbf{140.00}$\pm$\textbf{26.92} \\
            & Delivery &    86.88$\pm$9.49 &    83.12$\pm$7.26 &    96.25$\pm$7.81 &  \textbf{103.75}$\pm$\textbf{10.53} \\
\cmidrule{1-6}
\multirow{2}{*}{Asymmetric Advantages} & Onion Placement \& Delivery (Pot1) &  233.13$\pm$17.75 &  256.25$\pm$18.66 &  \textbf{282.88}$\pm$\textbf{17.03} &  260.63$\pm$18.36 \\
            & Delivery (Pot2) &  215.00$\pm$16.58 &  250.00$\pm$19.36 &  258.13$\pm$21.71 &  \textbf{268.00}$\pm$\textbf{9.79}  \\
\cmidrule{1-6}
\multirow{2}{*}{Soup Coordination} & Onion Placement \& Delivery &   199.38$\pm$6.09 &  105.00 $\pm$ 32.78 &   198.75$\pm$4.84 &   \textbf{219.38}$\pm$\textbf{3.47} \\
            & Tomato Place \& Delivery &   44.38$\pm$29.04 &   192.50$\pm$9.68 &  128.12$\pm$30.76 &   \textbf{220.63}$\pm$\textbf{3.47} \\
\cmidrule{1-6}
\multirow{2}{*}{Distant Tomato} & Tomato Placement &   38.75$\pm$30.79 &   27.50$\pm$27.27 &  148.75$\pm$68.36 &   \textbf{210.00}$\pm$\textbf{15.00} \\
            & Tomato Place \& Delivery &  175.62$\pm$24.35 &  180.00$\pm$22.36 &  198.12$\pm$37.20 &  \textbf{251.25}$\pm$\textbf{23.41} \\
\cmidrule{1-6}
\multirow{2}{*}{Many Orders} & Tomato Placement &  140.62$\pm$32.59 &  170.00$\pm$33.91 &  248.75$\pm$29.55 &  \textbf{256.36}$\pm$\textbf{35.99} \\
            & Delivery &  194.38$\pm$12.48 &  175.63$\pm$35.61 &  208.13$\pm$25.42 &  \textbf{241.21}$\pm$\textbf{12.97} \\
\bottomrule
\end{tabular}
} 
\end{table}

For fair comparison, the training steps and the pool sizes of FCP, MEP and HSP are all set the same, while our proposed approach HAPLAN is based on LLMs without the need of training one extra coordination policy. The results in Table~\ref{tab:exp1} demonstrates that HAPLAN achieves the highest scores across almost all scenarios. For example, in the layout of \textit{Distant Tomato}, when coordinating with the partner that prefers to place tomatoes in the pot, HAPLAN obtains score several times higher than FCP and MEP, as well as achieves a performance improvement of over 40\% compared to HSP. This indicates that methods like FCP and MEP, which train on a pool of partners, struggle to capture specific partner behavior preferences during testing, resulting in a lack of adaptive coordination. HSP is relatively better than them as it explicitly models the human biases. However, our approach still achieves superior performance to HSP, indicating that including LLMs allows better adaptation to various types of partners for improved coordination.


\subsubsection{Coordinating with Real Humans}

In addition to the scripted proxies, we also conducted experiments with real human participants to evaluate the effectiveness of different methods in real human-AI coordination scenarios. Compared to the scripted agents, human players are more flexible and dynamic, making coordination with real human more challenging. In specific, we involve a total of $12$ volunteers in the experiment, each of whom had limited prior experience with the \textit{Overcooked-AI} game before. To test a method on one specific layout, we allow the human player to have three rounds of coordination with the AI agent, allowing us to observe the changes in coordination baselines. Unlike other baseline algorithms, when testing our method, we allow the human partner to engage in natural language communication with the AI agent before the start of each coordination round.
The final results are shown in Table~\ref{tab:exp2}.

\begin{table}[t!]
\centering
\caption{Experimental results on Overcooked-AI environment of HAPLAN and baselines when coordinating with real humans. The best values in each round of coordination have been \textbf{bolded}.}
\label{tab:exp2}
\centering
\resizebox{\linewidth}{!}{
\begin{tabular}{c|c|c|c|c|c|c}
\toprule
                   &         &    Counter Circle & Asymmetric Advantages & Soup Coordination &    Distant Tomato &       Many Orders \\
\midrule
\multirow{4}{*}{\makecell{First Round}} & FCP &  120.00$\pm$12.64 &      336.00$\pm$24.97 &  \textbf{192.00}$\pm$\textbf{20.39} &  314.00$\pm$25.37 &  329.00$\pm$32.38 \\
                   & MEP &  \textbf{140.00}$\pm$\textbf{21.91} &      346.00$\pm$25.37 &  184.00$\pm$14.96 &  310.00$\pm$22.36 &  318.00$\pm$31.55 \\
                   & HSP &  \textbf{140.00}$\pm$\textbf{15.49} &      \textbf{350.00}$\pm$\textbf{34.92} &   184.00$\pm$8.01 &  \textbf{330.00}$\pm$\textbf{24.08} &  340.00$\pm$43.81 \\
                   & HAPLAN &  138.00$\pm$20.88 &      338.00$\pm$27.49 &  \textbf{192.00}$\pm$\textbf{18.33} &  324.00$\pm$29.39 &  \textbf{349.00}$\pm$\textbf{63.01} \\
\cmidrule{1-7}
\multirow{4}{*}{\makecell{Second Round}} & FCP &  138.00$\pm$10.77 &      350.00$\pm$18.43 &  194.00$\pm$18.01 &  338.00$\pm$18.86 &  340.00$\pm$29.66 \\
                   & MEP &  154.00$\pm$12.81 &      350.00$\pm$27.21 &  186.00$\pm$12.81 &  332.00$\pm$20.39 &  342.00$\pm$36.27 \\
                   & HSP &  154.00$\pm$15.62 &      \textbf{362.00}$\pm$\textbf{18.86} &  196.00$\pm$14.96 &  348.00$\pm$18.33 &  372.00$\pm$37.09 \\
                   & HAPLAN &  \textbf{160.00}$\pm$\textbf{15.49} &      360.00$\pm$25.29 &  \textbf{204.00}$\pm$\textbf{21.54} &  \textbf{356.00}$\pm$\textbf{17.43} &  \textbf{382.00}$\pm$\textbf{58.95} \\
\cmidrule{1-7}
\multirow{4}{*}{\makecell{Third Round}} & FCP &  136.00$\pm$17.43 &      350.00$\pm$25.69 &  198.00$\pm$28.91 &  336.00$\pm$34.41 &  349.00$\pm$23.01 \\
                   & MEP &  158.00$\pm$16.61 &      368.00$\pm$20.39 &  196.00$\pm$12.00 &  340.00$\pm$21.91 &  350.00$\pm$36.05 \\
                   & HSP &  160.00$\pm$12.64 &      368.00$\pm$27.12 &  198.00$\pm$10.77 &  352.00$\pm$25.61 &  376.00$\pm$33.22 \\
                   & HAPLAN &  \textbf{168.00}$\pm$\textbf{13.26} &      \textbf{384.00}$\pm$\textbf{21.54} &  \textbf{214.00}$\pm$\textbf{15.62} &  \textbf{370.00}$\pm$\textbf{22.36} &  \textbf{414.00}$\pm$\textbf{56.61} \\
\bottomrule
\end{tabular}
}
\end{table}

From the experimental results, we can mainly conclude two points: 
1) Firstly, under the same number of rounds, our method generally achieves better coordination performance with the human partner.
2) Secondly, our method exhibits a more significant performance improvement through the three rounds of human-AI coordination. 
In specific, our approach consistently outperforms the baseline algorithms across all layouts after the second round and on some specific layouts our approach attains the best coordination performance right from the first round. For example, on the \textit{Asymmetric Advantages} layout, HAPLAN's performance is only slightly below that of HSP in the first round, and starting from the second round, HAPLAN can achieve the highest coordination score with human partner on this layout. This reveals that on one hand the inclusion of LLMs can facilitate the AI agent's coordination with real human partner; on the other hand, LLMs make the AI's behavior more interpretable, helping the humans become familiar with and adapt to the task more quickly. A deeper discussion about why LLMs bring about such gains is provided in the next section.

\subsection{Analysis of LLMs in Human-AI Coordination}
\label{exp:analysis}

In fact, the previous traditional methods to some extent separate AI from humans, leaving AI agent an incomprehensible black box for human. 
The inclusion of LLMs strengthens the interaction between human and AI, allowing both human and AI to understand and benefit each other. In this section, we show the role of LLMs in human-AI coordination, and analyse why LLMs can enhance the coordination performance.

 \begin{wrapfigure}{r}{0.74\linewidth}
    \centering
    \includegraphics[width=\linewidth]{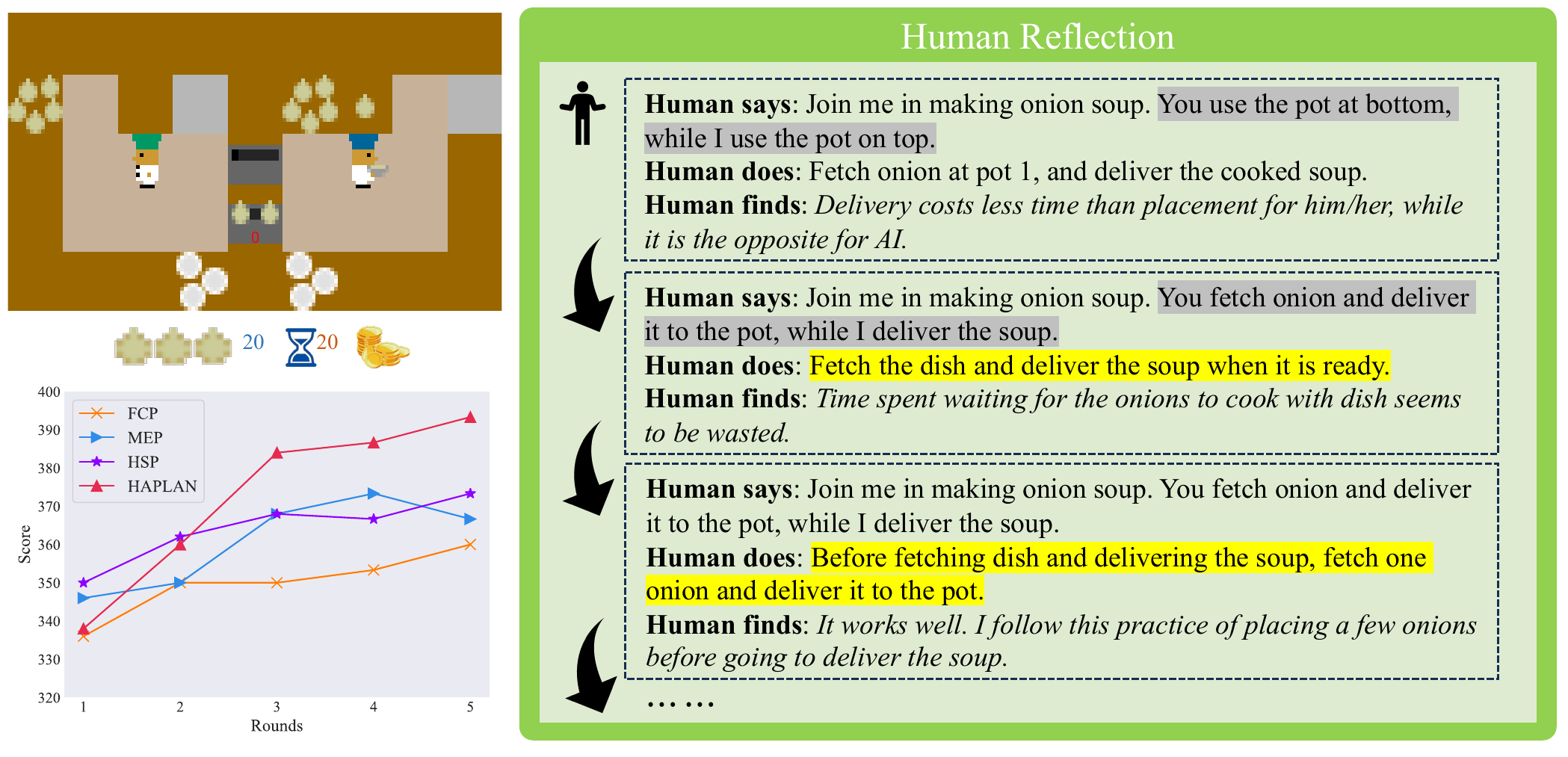}
    \caption{Details of results on the \textit{Asymmetric Advantages} layout.}
    \label{fig:ai2human}
 \end{wrapfigure}
\paragraph{AI to Human: Explainable AI behaviors} In~\ref{fig:ai2human}, we let volunteers conduct $5$ rounds of tests on the \textit{Asymmetric Advantages} layout, where the results show that our method obtains the fastest score improvement. Besides, we also provide an example to explain the details in this process.
This case reveals that when the AI agent adopts predictable behavioral actions, the human participant can quickly familiarize themselves with the task by trying his/her ideas, and gradually figure out strategies that can be effectively deployed alongside AI. In contrast, for the traditional methods, though the human participant becomes more familiar with the task, it remains challenging to discern how to coordinate with the AI agent since their behavior is difficult to comprehend and unexpected.

\begin{figure}[htbp]
    \centering
    \includegraphics[width=0.9\linewidth]{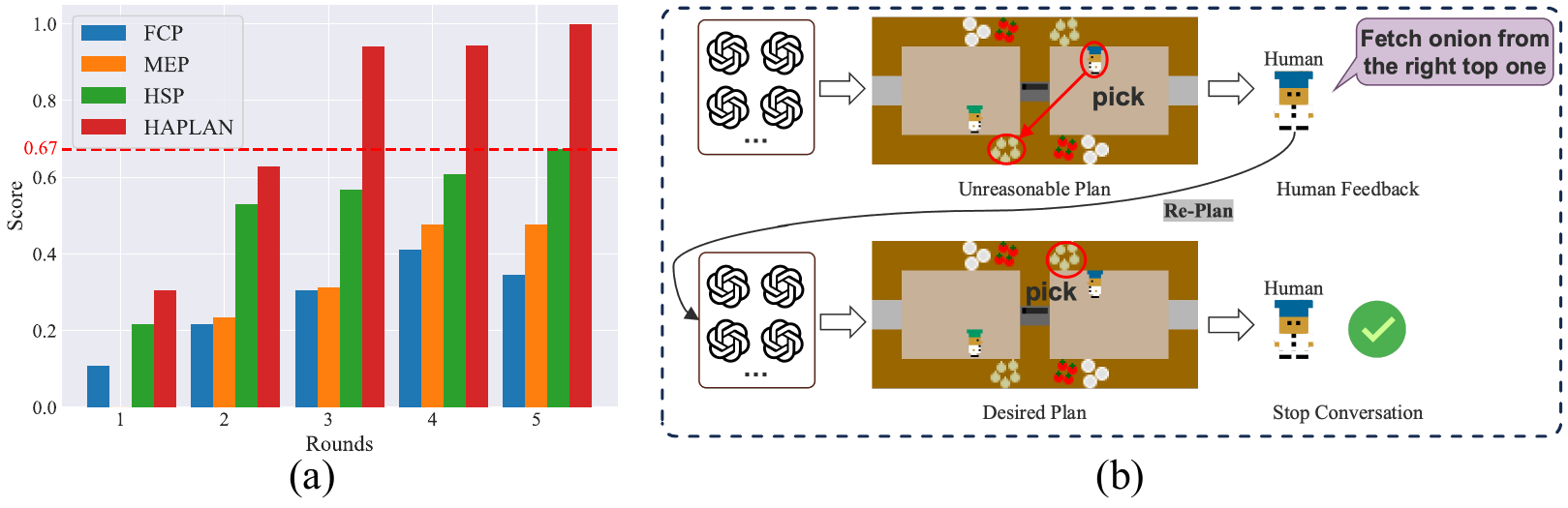}
    \caption{(a) Normalized scores on \textit{Many Orders}. (b) An example of human-AI conversation.}
    \label{fig:analysis1}
\end{figure}

\paragraph{Human to AI: Incorporating human domain knowledge} On the other hand, the utilization of LLMs can help incorporate the human partner's domain knowledge into the human-AI coordination, 
which can help discover some coordination patterns that are challenging for traditional learning methods to explore. In some complex scenarios, this can significantly contribute to achieving a higher level of human-AI coordination performance. Taking \textit{Many Orders} layout as an example, humans intuitively tend to believe that actively utilizing all three pots is essential for completing the task efficiently. With this insight, as shown in~\ref{fig:analysis1}(a),
our method can achieve exceptionally high scores after the third round, significantly surpassing the highest score of other methods.
Thus, from this perspective, the inclusion of LLMs can help incorporate the human's domain knowledge into the coordination, enabling achieving near optimal performance even in some complex scenarios.


\begin{figure}[htbp]
    \centering
    \includegraphics[width=0.87\linewidth]{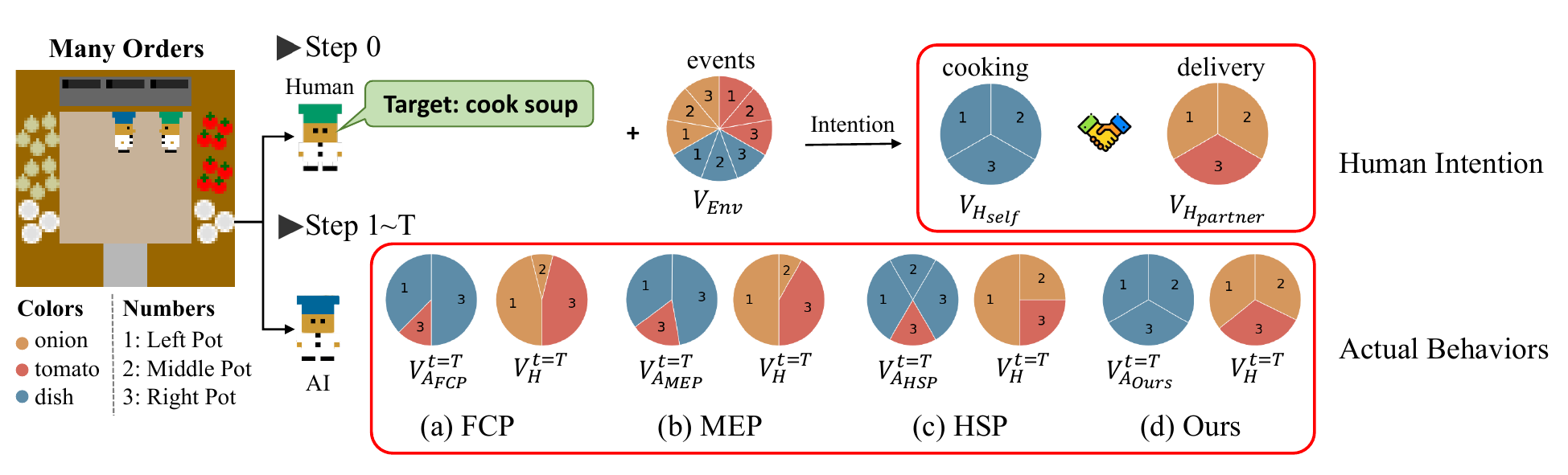}
    \caption{Overview of the human-AI value alignment. Colors denote task types and numbers indicate pot usage, e.g., the red sector of label 1 means placing onions to pot 1, the blue sector of label 2 means delivering the soup in pot 2. $V_{H_{self}}$ and $V_{H_{partner}}$ denote the human's initial intention regarding what they do respectively. Subsequent pie charts show actual event proportions post-trajectory.}
    \label{fig:analysis2}
\end{figure}

 \paragraph{Human-AI value alignment} Moreover, by introducing LLMs, our method can achieve human-centered human-AI value alignment. That is to say, through multiple rounds of human-AI 
 interactive dialogues, the AI agent can gain a comprehensive understanding of the human partner's thoughts and intentions, ensuring the consistency of the coordination behavior of the entire human-AI coordination system. Such as in the case shown in \ref{fig:analysis1}(b), though the human partner expects the AI agent to fetch onions, it is still possible that the AI agent generates unreasonable plan like fetching the onions in the bottom left corner. In such situation, the human partner can continue to correct the AI agent through dialogue, ensuring a desired plan for the AI agent. Similar conclusions can be observed in other layouts. To further provide a quantitative analysis, we present the value alignment results of different methods in the \textit{Many Orders} layout in~\ref{fig:analysis2}. The results indicate that our method exhibits behavior patterns closest to human value expectation. Totally, our method achieves better human-AI value alignment results, which holds significant value in ensuring the reliability and consistency of the entire human-AI coordination system.


\subsection{Ablation Study for Utilizing Multiple Sessions}
\label{exp:ablation}


Our method HAPLAN proposes to use multiple sessions of the LLM to make conventions, where we first decompose the entire problem into several sub-problems, and then utilize separate sessions to solve each sub-problem. To validate whether our practice of utilizing multiple sessions can enhance the reasoning capability, we compare with one baseline called Integrate-LLM (more details refer to Appendix~\ref{appx:integration_llm}) that only utilizes one single session. This baseline can be seen as an application that combines Chain-of-Thought (CoT) and Least-to-Most (L2M) on Overcooked-AI.
A specific point to note is that, both CoT and L2M also require the problem decomposition, so is Integrate-LLM.
Thus, for a fair comparison, we equally decompose the problem into $4$ sub-problems and design the same prompts for both Integrate-LLM and our method. Besides, to further validate how the problem decomposition affects the reasoning results, we have additionally included one comparison that decomposes the whole problem into $5$ sub-problems, denoted as HAPLAN-5. 
We provide the reasoning accuracy of all these three methods in~\ref{tab:multiple_llm_overcooked_ai_easy}.
From the results we can see that, our method consistently achieves higher accuracy than Integrate-LLM in all sub-problems as well as the final solution, which reveals that our approach of utilizing multiple sessions demonstrates superior reasoning capability on this task. Besides, HAPLAN-5 achieves even higher reasoning accuracy, indicating that appropriate problem decomposition also benefits the final reasoning quality. Note that since HAPLAN-5 performs better, the results in our main experiments are obtained by HAPLAN-5.

\begin{table}[htbp]
    \centering
    \caption{Reasoning accuracy on the Overcooked-AI environment. ``Subprob.'' denotes sub-problem. The preference of ``Placement: 1, Delivery: None'' means requiring to place ingredients in pot $1$ without delivery requirement. Here we only provide the results for \textit{Placement} task; the complete results can be found in Appendix~\ref{appx:reasoning_results_oc}.}
    \label{tab:multiple_llm_overcooked_ai_easy}
    \centering
    \resizebox{\linewidth}{!}{
    \begin{tabular}{c|c|c|c|c|c|c|c|c}
    \toprule
        Preference & Method & Subprob. 0 & Subprob. 1 & Subprob. 2 & Subprob. 3 & Subprob. 4 & Subprob. 5 & Final Solution \\ \midrule
       \multirow{3}{*}{\makecell{Placement: 1\\ Delivery: None}} & Integrate-LLM & / & 40\% & 100\% & 90\% & 50\% & / & 0\% \\
             & HAPLAN & / & 80\% & 100\% & 100\% & 60\% & / & 60\% \\
             & HAPLAN-5 & 100\% & 90\% & 100\% & 100\% & / & 100\% & 90\% \\ \cmidrule{1-9}
           \multirow{3}{*}{\makecell{Placement: 2+3\\ Delivery: None}} & Integrate-LLM & / & 30\% & 100\% & 100\% & 40\% & / & 0\% \\
             & HAPLAN & / & 80\% & 90\% & 80\% & 70\% & / & 60\% \\
             & HAPLAN-5 & 80\% & 80\% & 100\% & 100\% & / & 100\% & 80\% \\ \cmidrule{1-9}
           \multirow{3}{*}{\makecell{Placement: 1+2+3\\ Delivery: None}} & Integrate-LLM & / & 90\% & 90\% & 100\% & 100\% & / & 90\% \\
             & HAPLAN & / & 100\% & 100\% & 100\% & 100\% & / & 100\% \\
             & HAPLAN-5 & 100\% & 100\% & 100\% & 100\% & / & 100\% & 100\% \\ 
    \bottomrule
    \end{tabular}
}
\end{table}

 \begin{wraptable}{r}{0.45\linewidth}
    \centering
    \caption{Reasoning accuracy on the \textit{Symbolic Manipulation} benchmark.}
    \label{tab:multiple_llm_symbolic}
    \resizebox{\linewidth}{!}{
    \begin{tabular}{cccccc}
    \toprule
    \multirow{2}{*}{Method} & \multicolumn{5}{c}{Number of Words} \\ \cmidrule{2-6}
         & L=4 & L=6 & L=8 & L=10 & L=12 \\ \midrule
        Standard prompting  (original paper) & 0 & 0 & 0 & 0 & 0 \\ \midrule
        Chain-of-Thought  (original paper) & 84.2 & 69.2 & 50.2 & 39.8 & 31.8 \\ \midrule
        Least-to-Most  (original paper) & 94 & 88.4 & 83 & 76.4 & 74 \\ \midrule
        Least-to-Most (GPT-3.5) & \textbf{100} & \textbf{100} & 85 & 70 & 75 \\ \midrule
        Ours (2 Session) & \textbf{100} & \textbf{100} & \textbf{100} & \textbf{95} & \textbf{95} \\ 
    \bottomrule
    \end{tabular}
}
\end{wraptable}
\subsection{Additional Results on Reasoning Benchmarks}\label{exp:reasoning}
To validate the generality of our approach, we further conduct evaluation on several popular reasoning benchmarks. More details about these benchmarks and the complete experimental results can be found in Appendix~\ref{appx:reasoning_benchmark} and~\ref{appx:reasoning_results} respectively. Here, the results on the \textit{Symbolic Manipulation} benchmark are presented in~\ref{tab:multiple_llm_symbolic}. Besides the results reported in the original paper, we also reproduce the Least-to-Most method using the latest GPT-3.5 model and find that it exhibits some performance improvement compared to the results reported in its original paper. We hypothesize that this improvement comes from the update of GPT-3.5. Despite this, our method that utilizes $2$ sessions obtains the best performance in all cases with different numbers of words. Moreover, our method also exhibits minimal performance degradation when the number of words increases, still achieving an accuracy of up to 95\% which is more than 20\% higher than that of Least-to-Most. 
The results demonstrate the effectiveness of employing multiple sessions for reasoning tasks in various domains.

\section{Conclusion}

We propose HAPLAN, an efficient approach to making preparatory language-based conventions for human-AI coordination. To improve the reasoning abilities of LLMs, we propose to decompose a complex problem into several sub-problems and assign each of them to a new session sequentially. For a more efficient coordination, we propose to incorporate a human validation process to review the developed conventions. Experiments on the Overcooked-AI with human proxy models demonstrate the superiority of our approach compared with baselines. When coordinating real humans, our method also achieves higher performance with a better alignment to human preferences. Furthermore, we find that our idea can also be used to solve general reasoning tasks and show its effectiveness on benchmarks of symbolic manipulation, compositional reasoning and math reasoning.

\paragraph{Limitation and Future Work}
In our current implementation, problem decomposition is provided by humans. While we offer a general problem decomposition approach, specific problems may benefit from more efficient decomposition. One straightforward approach is to employ an additional session to suggest problem decomposition. Moreover, we are having humans review the conventions generated by the LLM. However, it is also possible to consider having the LLM itself review the content it generates. We plan to leave them for future works.

\bibliographystyle{unsrt}
\bibliography{sample-base}

\begin{thebibliography}{10}

\bibitem{overcooked-ai}
Micah Carroll, Rohin Shah, Mark~K. Ho, Tom Griffiths, Sanjit~A. Seshia, Pieter
  Abbeel, and Anca~D. Dragan.
\newblock On the utility of learning about humans for human-ai coordination.
\newblock In {\em Advances in Neural Information Processing Systems 32}, pages
  5175--5186, 2019.

\bibitem{goertzel2007artificial}
Ben Goertzel and Cassio Pennachin.
\newblock {\em Artificial General Intelligence}.
\newblock Springer, 2007.

\bibitem{DBLP:journals/chb/Endsley23}
Mica~R. Endsley.
\newblock Supporting human-ai teams: Transparency, explainability, and
  situation awareness.
\newblock {\em Computers in Human Behavior}, 140:107574, 2023.

\bibitem{instructRL}
Hengyuan Hu and Dorsa Sadigh.
\newblock Language instructed reinforcement learning for human-ai coordination.
\newblock In {\em International Conference on Machine Learning}, pages
  13584--13598, 2023.

\bibitem{DBLP:journals/corr/abs-2210-05125}
Hengyuan Hu, David~J. Wu, Adam Lerer, Jakob~N. Foerster, and Noam Brown.
\newblock Human-ai coordination via human-regularized search and learning.
\newblock {\em {arXiv Preprint arXiv:}2210.05125}, 2022.

\bibitem{parekh2023learning}
Sagar Parekh and Dylan~P. Losey.
\newblock Learning latent representations to co-adapt to humans.
\newblock {\em Autonomous Robots}, 47(6):771--796, 2023.

\bibitem{DBLP:conf/icml/HuLCPBF21}
Hengyuan Hu, Adam Lerer, Brandon Cui, Luis Pineda, Noam Brown, and Jakob~N.
  Foerster.
\newblock Off-belief learning.
\newblock In {\em {International Conference on Machine Learning}}, pages
  4369--4379, 2021.

\bibitem{DBLP:conf/nips/CuiHPF21}
Brandon Cui, Hengyuan Hu, Luis Pineda, and Jakob~N. Foerster.
\newblock K-level reasoning for zero-shot coordination in hanabi.
\newblock In {\em {Advances in Neural Information Processing Systems 34}},
  pages 8215--8228, 2021.

\bibitem{DBLP:conf/iclr/LaidlawD22}
Cassidy Laidlaw and Anca~D. Dragan.
\newblock The boltzmann policy distribution: Accounting for systematic
  suboptimality in human models.
\newblock In {\em {The Tenth International Conference on Learning
  Representations}}, 2022.

\bibitem{hsp}
Chao Yu, Jiaxuan Gao, Weilin Liu, Botian Xu, Hao Tang, Jiaqi Yang, Yu~Wang, and
  Yi~Wu.
\newblock Learning zero-shot cooperation with humans, assuming humans are
  biased.
\newblock In {\em {The Eleventh International Conference on Learning
  Representations}}, 2023.

\bibitem{jaderberg2017population}
Max Jaderberg, Valentin Dalibard, Simon Osindero, Wojciech~M Czarnecki, Jeff
  Donahue, Ali Razavi, Oriol Vinyals, Tim Green, Iain Dunning, Karen Simonyan,
  et~al.
\newblock Population based training of neural networks.
\newblock {\em {arXiv Preprint arXiv:1711.09846}}, 2017.

\bibitem{fcp}
Johannes Heinrich, Marc Lanctot, and David Silver.
\newblock Fictitious self-play in extensive-form games.
\newblock In {\em {International Conference on Machine Learning}}, pages
  805--813, 2015.

\bibitem{mep}
Rui Zhao, Jinming Song, Yufeng Yuan, Haifeng Hu, Yang Gao, Yi~Wu, Zhongqian
  Sun, and Wei Yang.
\newblock Maximum entropy population-based training for zero-shot human-ai
  coordination.
\newblock In {\em Thirty-Seventh {AAAI} Conference on Artificial Intelligence},
  pages 6145--6153, 2023.

\bibitem{DBLP:journals/cor/NourmohammadiFN22}
Amir Nourmohammadi, Masood Fathi, and Amos H.~C. Ng.
\newblock Balancing and scheduling assembly lines with human-robot
  collaboration tasks.
\newblock {\em Computers and Operations Research}, 140:105674, 2022.

\bibitem{DBLP:conf/hci/GleichaufSW22}
Katharina Gleichauf, Ramona Schmid, and Verena Wagner{-}Hartl.
\newblock Human-robot-collaboration in the healthcare environment: An
  exploratory study.
\newblock In {\em {the Twenty-Fourth International Conference on Human-Computer
  Interaction}}, pages 231--240, 2022.

\bibitem{DBLP:conf/nips/SiuPCZLPCA21}
Ho~Chit Siu, Jaime~Daniel Pe{\~{n}}a, Edenna Chen, Yutai Zhou, Victor~J. Lopez,
  Kyle Palko, Kimberlee~C. Chang, and Ross~E. Allen.
\newblock Evaluation of human-ai teams for learned and rule-based agents in
  hanabi.
\newblock In {\em {Advances in Neural Information Processing Systems 34}},
  pages 16183--16195, 2021.

\bibitem{strouse2021collaborating}
DJ~Strouse, Kevin McKee, Matt Botvinick, Edward Hughes, and Richard Everett.
\newblock Collaborating with humans without human data.
\newblock In {\em {Advances in Neural Information Processing Systems 34}},
  pages 14502--14515, 2021.

\bibitem{thatcher2021foundations}
Robert~W Thatcher and E~Roy John.
\newblock {\em Foundations of cognitive processes}.
\newblock Routledge, 2021.

\bibitem{sutton2018reinforcement}
Richard~S Sutton and Andrew~G Barto.
\newblock {\em Reinforcement Learning: An Introduction}.
\newblock MIT press, 2018.

\bibitem{xue2022heterogeneous}
Ke~Xue, Yutong Wang, Lei Yuan, Cong Guan, Chao Qian, and Yang Yu.
\newblock Heterogeneous multi-agent zero-shot coordination by coevolution.
\newblock {\em {arXiv Preprint arXiv:2208.04957}}, 2022.

\bibitem{charakorn2022generating}
Rujikorn Charakorn, Poramate Manoonpong, and Nat Dilokthanakul.
\newblock Generating diverse cooperative agents by learning incompatible
  policies.
\newblock In {\em {The Eleventh International Conference on Learning
  Representations}}, 2022.

\bibitem{fastap}
Ziqian Zhang, Lei Yuan, Lihe Li, Ke~Xue, Chengxing Jia, Cong Guan, Chao Qian,
  and Yang Yu.
\newblock Fast teammate adaptation in the presence of sudden policy change.
\newblock In {\em {Uncertainty in Artificial Intelligence}}, pages 2465--2476,
  2023.

\bibitem{shih2020critical}
Andy Shih, Arjun Sawhney, Jovana Kondic, Stefano Ermon, and Dorsa Sadigh.
\newblock On the critical role of conventions in adaptive human-ai
  collaboration.
\newblock In {\em {The Ninth International Conference on Learning
  Representations}}, 2020.

\bibitem{gordon2023impact}
John-Stewart Gordon.
\newblock {\em The Impact of Artificial Intelligence on Human Rights
  Legislation: A Plea for an AI Convention}.
\newblock Springer Nature, 2023.

\bibitem{weng2020odqa}
Lilian Weng.
\newblock How to build an open-domain question answering system?
\newblock {\em lilianweng.github.io}, Oct 2020.

\bibitem{survey}
Wayne~Xin Zhao, Kun Zhou, Junyi Li, Tianyi Tang, Xiaolei Wang, Yupeng Hou,
  Yingqian Min, Beichen Zhang, Junjie Zhang, Zican Dong, Yifan Du, Chen Yang,
  Yushuo Chen, Zhipeng Chen, Jinhao Jiang, Ruiyang Ren, Yifan Li, Xinyu Tang,
  Zikang Liu, Peiyu Liu, Jian{-}Yun Nie, and Ji{-}Rong Wen.
\newblock A survey of large language models.
\newblock {\em {arXiv Preprint arXiv}:2303.18223}, 2023.

\bibitem{DBLP:journals/corr/abs-2303-00855}
Wenlong Huang, Fei Xia, Dhruv Shah, Danny Driess, Andy Zeng, Yao Lu, Pete
  Florence, Igor Mordatch, Sergey Levine, Karol Hausman, and Brian Ichter.
\newblock Grounded decoding: Guiding text generation with grounded models for
  robot control.
\newblock {\em {arXiv Preprint arXiv}:2303.00855}, 2023.

\bibitem{DBLP:conf/acl/QiaoO0CYDTHC23}
Shuofei Qiao, Yixin Ou, Ningyu Zhang, Xiang Chen, Yunzhi Yao, Shumin Deng,
  Chuanqi Tan, Fei Huang, and Huajun Chen.
\newblock Reasoning with language model prompting: {A} survey.
\newblock In {\em {The Sixty-First Annual Meeting of the Association for
  Computational Linguistics}}, pages 5368--5393, 2023.

\bibitem{DBLP:conf/acl/LeeHPP023}
Gibbeum Lee, Volker Hartmann, Jongho Park, Dimitris Papailiopoulos, and
  Kangwook Lee.
\newblock Prompted llms as chatbot modules for long open-domain conversation.
\newblock In {\em {The Sixty-First Annual Meeting of the Association for
  Computational Linguistics}}, pages 4536--4554, 2023.

\bibitem{chain-of-thought}
Jason Wei, Xuezhi Wang, Dale Schuurmans, Maarten Bosma, Brian Ichter, Fei Xia,
  Ed~H. Chi, Quoc~V. Le, and Denny Zhou.
\newblock Chain-of-thought prompting elicits reasoning in large language
  models.
\newblock In {\em {Advances in Neural Information Processing Systems 35}},
  2022.

\bibitem{least-to-most}
Denny Zhou, Nathanael Sch{\"{a}}rli, Le~Hou, Jason Wei, Nathan Scales, Xuezhi
  Wang, Dale Schuurmans, Claire Cui, Olivier Bousquet, Quoc~V. Le, and Ed~H.
  Chi.
\newblock Least-to-most prompting enables complex reasoning in large language
  models.
\newblock In {\em {The Eleventh International Conference on Learning
  Representations}}, 2023.

\bibitem{hallucination}
Yue Zhang, Yafu Li, Leyang Cui, Deng Cai, Lemao Liu, Tingchen Fu, Xinting
  Huang, Enbo Zhao, Yu~Zhang, Yulong Chen, Longyue Wang, Anh~Tuan Luu, Wei Bi,
  Freda Shi, and Shuming Shi.
\newblock Siren's song in the {AI} ocean: {A} survey on hallucination in large
  language models.
\newblock {\em {arXiv Preprint arXiv}:2309.01219}, 2023.

\bibitem{yuan2022situ}
Luyao Yuan, Xiaofeng Gao, Zilong Zheng, Mark Edmonds, Ying~Nian Wu, Federico
  Rossano, Hongjing Lu, Yixin Zhu, and Song-Chun Zhu.
\newblock In situ bidirectional human-robot value alignment.
\newblock {\em Science robotics}, 7(68):eabm4183, 2022.

\bibitem{rmix}
Wei Qiu, Xinrun Wang, Runsheng Yu, Rundong Wang, Xu~He, Bo~An, Svetlana
  Obraztsova, and Zinovi Rabinovich.
\newblock {RMIX:} learning risk-sensitive policies for cooperative
  reinforcement learning agents.
\newblock In {\em {Advances in Neural Information Processing Systems 35}},
  pages 23049--23062, 2021.

\bibitem{DBLP:journals/corr/abs-2211-09935}
Shreyas~Sundara Raman, Vanya Cohen, Eric Rosen, Ifrah Idrees, David Paulius,
  and Stefanie Tellex.
\newblock Planning with large language models via corrective re-prompting.
\newblock {\em {arXiv Preprint arXiv}:2211.09935}, 2022.

\bibitem{DBLP:conf/icml/HuangAPM22}
Wenlong Huang, Pieter Abbeel, Deepak Pathak, and Igor Mordatch.
\newblock Language models as zero-shot planners: Extracting actionable
  knowledge for embodied agents.
\newblock In {\em {International Conference on Machine Learning}}, pages
  9118--9147, 2022.

\bibitem{sama}
Wenhao Li, Dan Qiao, Baoxiang Wang, Xiangfeng Wang, Bo~Jin, and Hongyuan Zha.
\newblock Semantically aligned task decomposition in multi-agent reinforcement
  learning.
\newblock {\em {arXiv Preprint arXiv}:2305.10865}, 2023.

\bibitem{proagent}
Ceyao Zhang, Kaijie Yang, Siyi Hu, Zihao Wang, Guanghe Li, Yihang Sun, Cheng
  Zhang, Zhaowei Zhang, Anji Liu, Song{-}Chun Zhu, Xiaojun Chang, Junge Zhang,
  Feng Yin, Yitao Liang, and Yaodong Yang.
\newblock Proagent: Building proactive cooperative {AI} with large language
  models.
\newblock {\em {arXiv Preprint arXiv}:2308.11339}, 2023.

\bibitem{llmagent}
Lei Wang, Chen Ma, Xueyang Feng, Zeyu Zhang, Hao Yang, Jingsen Zhang, Zhiyuan
  Chen, Jiakai Tang, Xu~Chen, Yankai Lin, Wayne~Xin Zhao, Zhewei Wei, and
  Ji{-}Rong Wen.
\newblock A survey on large language model based autonomous agents.
\newblock {\em {arXiv Preprint arXiv}:2308.11432}, 2023.

\bibitem{surveyfd}
Zhiheng Xi, Wenxiang Chen, Xin Guo, Wei He, Yiwen Ding, Boyang Hong, Ming
  Zhang, Junzhe Wang, Senjie Jin, Enyu Zhou, Rui Zheng, Xiaoran Fan, Xiao Wang,
  Limao Xiong, Yuhao Zhou, Weiran Wang, Changhao Jiang, Yicheng Zou, Xiangyang
  Liu, Zhangyue Yin, Shihan Dou, Rongxiang Weng, Wensen Cheng, Qi~Zhang,
  Wenjuan Qin, Yongyan Zheng, Xipeng Qiu, Xuanjing Huan, and Tao Gui.
\newblock The rise and potential of large language model based agents: {A}
  survey.
\newblock {\em {arXiv Preprint arXiv}:2309.07864}, 2023.

\bibitem{littman1994markov}
Michael~L Littman.
\newblock Markov games as a framework for multi-agent reinforcement learning.
\newblock In {\em Machine Learning}, pages 157--163. Elsevier, 1994.

\bibitem{roleplay}
Murray Shanahan, Kyle McDonell, and Laria Reynolds.
\newblock Role-play with large language models.
\newblock {\em {arXiv Preprint arXiv}:2305.16367}, 2023.

\bibitem{chatgpt}
{OpenAI}.
\newblock Introducing {ChatGPT}, 2022.

\bibitem{saycan}
Brian Ichter, Anthony Brohan, Yevgen Chebotar, Chelsea Finn, Karol Hausman,
  Alexander Herzog, Daniel Ho, Julian Ibarz, Alex Irpan, Eric Jang, Ryan
  Julian, Dmitry Kalashnikov, Sergey Levine, Yao Lu, Carolina Parada, Kanishka
  Rao, Pierre Sermanet, Alexander Toshev, Vincent Vanhoucke, Fei Xia, Ted Xiao,
  Peng Xu, Mengyuan Yan, Noah Brown, Michael Ahn, Omar Cortes, Nicolas Sievers,
  Clayton Tan, Sichun Xu, Diego Reyes, Jarek Rettinghouse, Jornell Quiambao,
  Peter Pastor, Linda Luu, Kuang{-}Huei Lee, Yuheng Kuang, Sally Jesmonth,
  Nikhil~J. Joshi, Kyle Jeffrey, Rosario~Jauregui Ruano, Jasmine Hsu, Keerthana
  Gopalakrishnan, Byron David, Andy Zeng, and Chuyuan~Kelly Fu.
\newblock Do as {I} can, not as {I} say: Grounding language in robotic
  affordances.
\newblock In {\em {Conference on Robot Learning}}, pages 287--318, 2022.

\bibitem{bc}
Dean Pomerleau.
\newblock Efficient training of artificial neural networks for autonomous
  navigation.
\newblock {\em {Neural Computation}}, 3(1):88--97, 1991.

\bibitem{scan}
Brenden~M. Lake and Marco Baroni.
\newblock Generalization without systematicity: On the compositional skills of
  sequence-to-sequence recurrent networks.
\newblock In {\em {International Conference on Machine Learning}}, pages
  2879--2888, 2018.

\bibitem{gsm}
Karl Cobbe, Vineet Kosaraju, Mohammad Bavarian, Mark Chen, Heewoo Jun, Lukasz
  Kaiser, Matthias Plappert, Jerry Tworek, Jacob Hilton, Reiichiro Nakano,
  Christopher Hesse, and John Schulman.
\newblock Training verifiers to solve math word problems.
\newblock {\em {arXiv Preprint arXiv}:2110.14168}, 2021.

\bibitem{drop}
Dheeru Dua, Yizhong Wang, Pradeep Dasigi, Gabriel Stanovsky, Sameer Singh, and
  Matt Gardner.
\newblock {DROP:} {A} reading comprehension benchmark requiring discrete
  reasoning over paragraphs.
\newblock In {\em The 2019 Conference of the North American Chapter of the
  Association for Computational Linguistics: Human Language Technologies},
  pages 2368--2378, 2019.

\end{thebibliography}

\appendix

\clearpage
\section{Appendix}
\subsection{Environment Details}
\label{appx:env}
\subsubsection{Overcooked-AI Environment}
\label{appx:overcooked_ai_env}

\begin{figure}[htbp]
    \centering
    \includegraphics[width=0.9\linewidth]{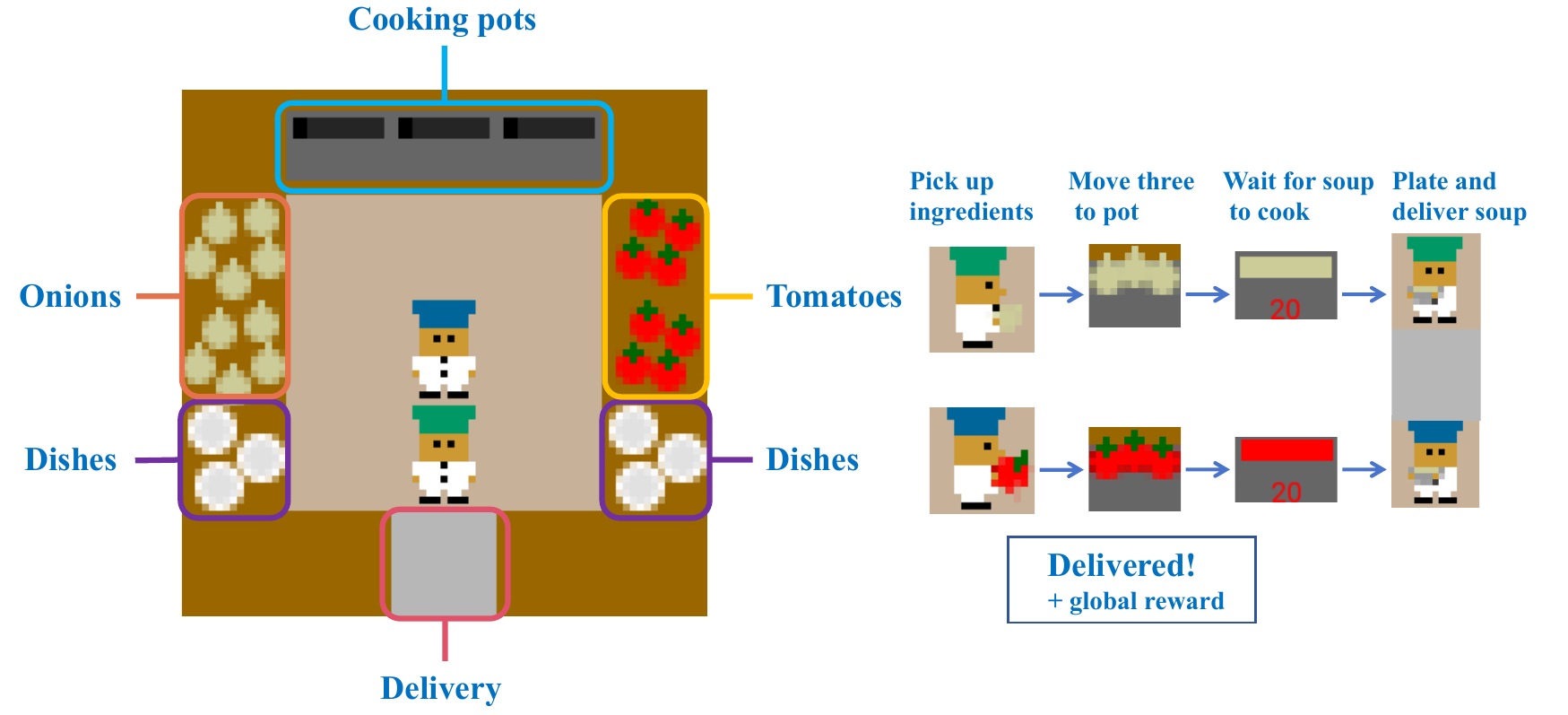} 
    \caption{Overcooked-AI~\cite{overcooked-ai} environment.}
    \label{fig:overcooked_env}
\end{figure}

Overcooked-AI~\cite{overcooked-ai} is a recent popular benchmark environment for human-AI coordination performance. In this environment, the goal of two players is to complete as many orders as possible within limited time, where each order corresponds to delivering a soup. In specific, each soup requires fetching ingredients  (e.g., onions, tomatoes), placing them in the pot, waiting the soup to cook, then picking up the soup and delivering it. The players must coordinate well to complete each step efficiently. Below, we concretely introduce the five layouts used in our experiments.

\begin{figure}[htbp]
    \centering
    \includegraphics[width=\linewidth]{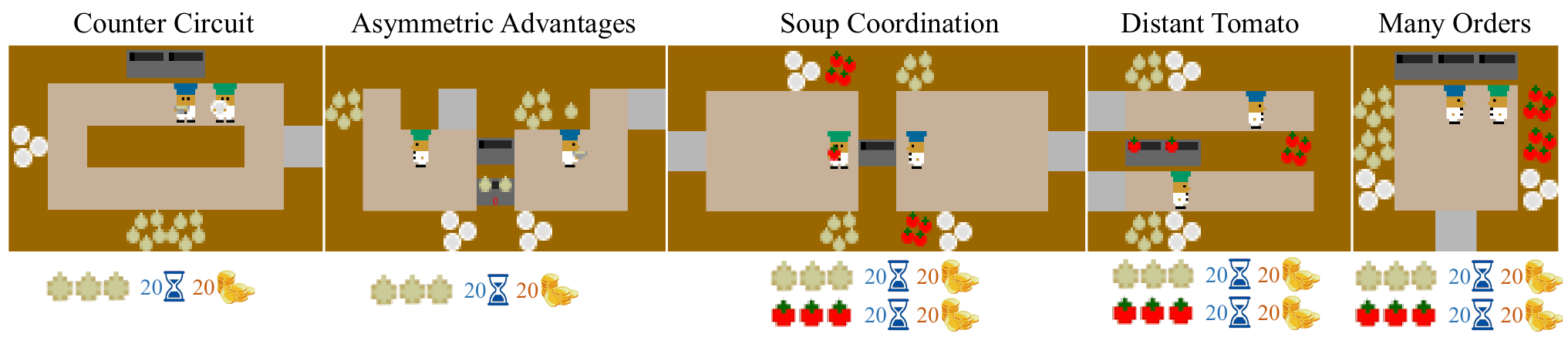} 
    \caption{Five layouts of the Overcooked-AI~\cite{overcooked-ai} environments used in our experiment. Respectively: \textit{Counter Circle}, \textit{Asymmetric Advantages}, \textit{Soup Coordination}, \textit{Distant Tomato} and \textit{Many Orders}.}
    \label{fig:layouts}
\end{figure}

\paragraph{Counter Circle} In the \textit{Counter Circle} layout, two players are placed in the same kitchen. There is a long, narrow counter stretches through the center of the kitchen, necessitating seamless coordination between players to prevent obstruction. Onions, dishes, and pots are positioned at the bottom, left, and top of the kitchen, respectively. Players must employ them appropriately for cooking and delivering the soup. Besides, two pots are located at the top, demanding effective utilization to enhance task completion efficiency.

\paragraph{Asymmetric Advantages} The \textit{Asymmetric Advantages} layout places two players in two separate kitchens characterized by strong asymmetry. In the left kitchen, it takes more time to fetch the onions as the onions are placed far away from the pot, while the server is close, making delivering convenient. In contrast, in the right kitchen, the situation is reversed: delivery becomes inconvenient, yet fetching the onions is easy. Thus, players can achieve high efficiency through effective coordination.

\paragraph{Soup Coordination} In the \textit{Soup Coordination} layout, players are also situated into two separate kitchens, yet these two kitchens are essentially symmetrical. Both kitchens contain tomatoes, onions, dishes, and a server, but only the area between the two kitchens features a pot, requiring players to coordinate effectively to complete orders.

\paragraph{Distant Tomato} In the \textit{Distant Tomato} layout, the two players similarly find them in two separate kitchens. In each kitchen, the onions, dishes, and pots are conveniently close, while the tomatoes are situated at a distance. Furthermore, there are two pots between the two kitchens, requiring both players to coordinate effectively to improve order completion.

\paragraph{Many Orders} The \textit{Many Orders} layout places players in the same kitchen, with onions and dishes on the left, tomatoes and dishes on the right, and three pots at the top. In this arrangement, players must coordinate effectively to ensure that all three pots are actively used, allowing for efficient soup cooking and delivery.

\subsubsection{Reasoning Benchmarks}
\label{appx:reasoning_benchmark}
These reasoning benchmarks are originally introduced in the paper of Least-to-Most~\cite{least-to-most}, including three types of tasks: \textit{Symbolic Manipulation}, \textit{Compositional Generalization} and \textit{Math Reasoning}. 

\paragraph{Symbolic Manipulation} The \textit{Symbolic Manipulation} task typically requires to concatenating the last latter of a series of words. In specific, the input for the LLM is a list of words and the corresponding expected output is the concatenation of the last letters of the words in the list. For example, for inputs ``listening, thinking, improve'' the corresponding output is ``gge'', since the last letters of the word list are respectively ``g'', ``g'' and ``e''.

\paragraph{Compositional Generalization} The \textit{Compositional Generalization} task utilizes SCAN~\cite{scan} as the benchmark, which typically requires mapping natural language command sentences to action sequences. For example, for command ``look thrice after jump'', the expected action sequence is ``JUMP LOOK LOOK LOOK''; for command ``run left and walk'', the expected action sequence is ``TURN\_LEFT RUN WALK''. 
 
\paragraph{Math Reasoning} \textit{Math Reasoning} task is aimed to test the reasoning capability of LLMs to solve math world problems in GSM8K~\cite{gsm} and DROP~\cite{drop}. One example question is ``Elsa has 5 apples, Anna has 2 more apples than Elsa. How many apples do they have together?''

\subsection{Baseline Details}\label{appx:baselines}
In our main experiments in~\ref{exp:main_exp}, we mainly compared our method with three popular human-AI coordination methods, respectively Fictitious Co-Play (FCP)~\cite{fcp}, Maximum Entropy Population-based training (MEP)~\cite{mep} and Hidden-utility Self-Play (HSP)~\cite{hsp}. The introduction of them are as follows:
\paragraph{FCP} Fictitious Co-Play (FCP)~\cite{fcp} is a two-stage approach to learn to collaborate with humans without human data. At the first stage, it builds a pool of partners which represent different conventions; while at the second stage, it trains a best-response agent to the obtained diverse partners and their checkpoints.
\paragraph{MEP} Maximum Entropy Population-based training   (MEP)~\cite{mep} also follows a two-stage framework, while it proposes to learn a diverse partner population through maximizing one centralized population entropy objective.
\paragraph{HSP} Hidden-utility Self-Play (HSP)~\cite{hsp} explicitly models the human biases as hidden reward functions. On this basis, it augments the policy pool with biased policies and afterwards trains an adaptive policy.

\subsection{Implementation Details}\label{sec:impl}
\subsubsection{Low-level Skill Learning}

\begin{figure}[htb]
    \centering
    \includegraphics[width=0.85\linewidth]{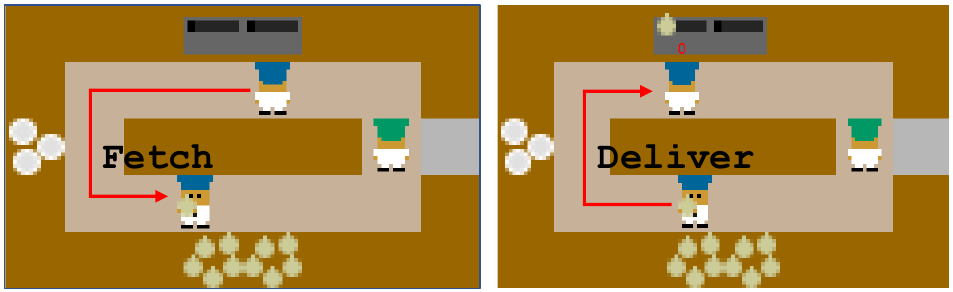}
    \caption{Visualization of the low-level skills, \texttt{Fetch}  (\textbf{left}) and \texttt{Deliver}  (\textbf{right}).}
    \label{fig:low-level}
\end{figure}

One significant component of our approach is the building of low-level skills. Only when the low-level skills are reliable, combined with the reasoning of the upper-level LLMs, can our AI agent coordinate effectively with humans. To achieve this goal, we propose a method that involves human players providing a small number of demonstrations, followed by learning lower-level skill policies through imitation learning techniques. This approach can be applied in scenarios where humans can provide a few demonstrations or where there exist a few demonstration segments (only fragmentary data required, \textbf{no need for complete coordination trajectory}). 

Specifically, in the Overcooked-AI environment, we have learned two types of low-level skills, as shown in~\ref{fig:low-level}, by requiring human players to collect a few demonstration trajectories. Each sub-task in this environment can be completed by this two two-level-skills. For example, the Delivery task (pick up the dish, scoop the soup and deliver the soup) in Overcooked-AI can be split into: \texttt{Fetch} the dish at \texttt{(x1,y1)}, \texttt{Deliver} the dish to \texttt{(x2,y2)}, \texttt{Fetch} the soup at \texttt{(x2,y2)}, \texttt{Deliver} the soup to \texttt{(x3,y3)}. The specific learning process can be divided into the following two steps:

\paragraph{Data Collection} To learn these two low-level skills, we firstly collected skill demonstration data by requiring the human participants to play a small number of games. In this process, to ensure the robustness of the learned skill policies when facing situations of different partner states, we had human players interact with one partner of random policy, thereby ensuring that the training data encounters various partner states. In our experiments, the demonstrations for training each skill are around $50$ trajectories.

\paragraph{Imitation Learn} After collecting the demonstration data, we began to learn the skill policies through behavior cloning method. In details, on one hand, we allow the skill parameters to be part of input into the skill policy. For example, for skill ``\texttt{Fetch} an onion at \texttt{(2,1)}'', we will concatenate the coordinate information \texttt{(2,1)} with the agent's observation and input them into the skill policy. On the other hand, we applied data augmentation to the demonstration data to further enhance the robustness of the learned skill policies. For example, for demonstration of fetching dish, if the pot is empty, we will augment new trajectory data with pot filled with ingredients. The reason is that whether there are ingredients in the pot does not affect the action of fetching the dish.

\subsubsection{Implementation of Integrate-LLM}
\label{appx:integration_llm}

In the ablation study of utilizing multiple sessions on Overcooked-AI, we have compared our method with one baseline called Integrate-LLM as mentioned in~\ref{exp:ablation}. Due to space constraints, we did not provide a detailed introduction to Integrate-LLM in the main text. Actually, it is implemented by simply replacing the utilization of multiple sessions in our method with using one single session, which means that this session will sequentially address all the sub-problems, and the relevant information of solving the sub-problems earlier will be retained as part of the prompt for obtaining the final reasoning results.
Besides, the prompt engineering employed by Integrate-LLMs is also nearly identical to our approach, which is listed in~\ref{sec:prompt}. 

This practice resembles Least-to-Most as they both utilize one single LLM (one session) to sequentially solve multiple sub-problems in order to solve the whole problem. The difference lies in that Least-to-Most directly applies LLM to decompose the problem, while Integrate-LLM employs a pre-defined problem decomposition scheme. Besides, it includes an example of problem solving that aligns with the problem decomposition scheme in the beginning of the prompt, which is similar to the practice of the Chain-of-Thought method. Thus, Integrate-LLM can be seen as an application that combines Chain-of-Thought and Least-to-Most on Overcooked-AI. The comparison with this baseline demonstrates that utilizing only one single session may struggle in some complex scenarios.

\subsection{Additional Results}

\begin{table}[t!]
    \centering
    \caption{Complete results of reasoning accuracy on the Overcooked-AI environment.}
    \label{tab:multiple_llm_overcooked_ai_complete}
    \centering
    \resizebox{\linewidth}{!}{
    \begin{tabular}{c|c|c|c|c|c|c|c|c|c}
    \toprule
        Rough preference & Specific preference & Method & Subprob. 0 & Subprob. 1 & Subprob. 2 & Subprob. 3 & Subprob. 4 & Subprob. 5 & Final Solution \\ \midrule
        \multirow{9}{*}{\makecell{Only Placement}} & \multirow{3}{*}{\makecell{Placement: 1\\ Delivery: None}} & Integrate-LLM & / & 40\% & 100\% & 90\% & 50\% & / & 0\% \\
          &   & HAPLAN & / & 80\% & 100\% & 100\% & 60\% & / & 60\% \\
          &   & HAPLAN-5 & 100\% & 90\% & 100\% & 100\% & / & 100\% & 90\% \\ \cmidrule{2-10}
          & \multirow{3}{*}{\makecell{Placement: 2+3\\ Delivery: None}} & Integrate-LLM & / & 30\% & 100\% & 100\% & 40\% & / & 0\% \\
          &   & HAPLAN & / & 80\% & 90\% & 80\% & 70\% & / & 60\% \\
          &   & HAPLAN-5 & 80\% & 80\% & 100\% & 100\% & / & 100\% & 80\% \\ \cmidrule{2-10}
          & \multirow{3}{*}{\makecell{Placement: 1+2+3\\ Delivery: None}} & Integrate-LLM & / & 90\% & 90\% & 100\% & 100\% & / & 90\% \\
          &   & HAPLAN & / & 100\% & 100\% & 100\% & 100\% & / & 100\% \\
          &   & HAPLAN-5 & 100\% & 100\% & 100\% & 100\% & / & 100\% & 100\% \\ \cmidrule{1-10}
        \multirow{9}{*}{Only Delivery} & \multirow{3}{*}{\makecell{Placement: None\\ Delivery: 2}} & Integrate-LLM & / & 40\% & 90\% & 80\% & 20\% & / & 0\% \\ 
        ~ & ~ & HAPLAN & / & 70\% & 100\% & 100\% & 100\% & / & 70\% \\ 
        ~ & ~ & HAPLAN-5 & 70\% & 70\% & 100\% & 100\% & / & 100\% & 70\% \\ \cmidrule{2-10}
        ~ & \multirow{3}{*}{\makecell{Placement: None\\ Delivery: 1+3}} & Integrate-LLM & / & 40\% & 100\% & 80\% & 30\% & / & 0\% \\ 
        ~ & ~ & HAPLAN & / & 60\% & 100\% & 100\% & 100\% & / & 60\% \\ 
        ~ & ~ & HAPLAN-5 & 80\% & 100\% & 100\% & 100\% & / & 100\% & 80\% \\ \cmidrule{2-10}
        ~ & \multirow{3}{*}{\makecell{Placement: None\\ Delivery: 1+2+3}} & Integrate-LLM & / & 80\% & 100\% & 90\% & 60\% & / & 70\% \\ 
        ~ & ~ & HAPLAN & / & 100\% & 100\% & 100\% & 100\% & / & 100\% \\ 
        ~ & ~ & HAPLAN-5 & 100\% & 100\% & 100\% & 100\% & / & 100\% & 100\% \\ \cmidrule{1-10}
        \multirow{12}{*}{Place \& Delivery} & \multirow{3}{*}{\makecell{Placement: 1\\ Delivery: 1}} & Integrate-LLM & / & 90\% & 100\% & 100\% & 100\% & / & 100\% \\ 
        ~ & ~ & HAPLAN & / & 100\% & 100\% & 100\% & 100\% & / & 100\% \\ 
        ~ & ~ & HAPLAN-5 & 100\% & 100\% & 100\% & 100\% & / & 100\% & 100\% \\ \cmidrule{2-10}
        ~ & \multirow{3}{*}{\makecell{Placement: 1\\ Delivery: 1+2}} & Integrate-LLM & / & 90\% & 100\% & 90\% & 70\% & / & 60\% \\ 
        ~ & ~ & HAPLAN & / & 100\% & 100\% & 100\% & 100\% & / & 100\% \\ 
        ~ & ~ & HAPLAN-5 & 100\% & 100\% & 100\% & 100\% & / & 100\% & 100\% \\ \cmidrule{2-10}
        ~ & \multirow{3}{*}{\makecell{Placement: 2+3\\ Delivery: 3}} & Integrate-LLM & / & 70\% & 100\% & 100\% & 90\% & / & 70\% \\ 
        ~ & ~ & HAPLAN & / & 80\% & 100\% & 100\% & 100\% & / & 80\% \\ 
        ~ & ~ & HAPLAN-5 & 100\% & 100\% & 100\% & 100\% & / & 100\% & 100\% \\ \cmidrule{2-10}
        ~ & \multirow{3}{*}{\makecell{Placement: 1\\ Delivery: 1+2+3}} & Integrate-LLM & / & 90\% & 100\% & 90\% & 90\% & / & 90\% \\ 
        ~ & ~ & HAPLAN & / & 100\% & 100\% & 100\% & 100\% & / & 100\% \\
        ~ & ~ & HAPLAN-5 & 100\% & 100\% & 100\% & 100\% & / & 100\% & 100\% \\ 
    \bottomrule
    \end{tabular}
}
\end{table}

\subsubsection{Complete Results of Reasoning Accuracy on Overcooked-AI}
\label{appx:reasoning_results_oc}

To validate whether utilizing multiple sessions of the LLM can enhance the reasoning capability, we conduct ablation to evaluate the reasoning accuracy on Overcooked-AI environment. Specifically, we have multiple tasks where each has certain preferences. We first generate several corresponding commands for each task, and subsequently we evaluate the accuracy of different methods in deducing planning results that fulfill these requirements through reasoning. We have provided some results in~\ref{exp:ablation}, while the complete results are presented in~\ref{tab:multiple_llm_overcooked_ai_complete}. It can be observed that our method consistently achieves the best reasoning results in this experiment.

\subsubsection{Complete Results on Reasoning Benchmarks}
\label{appx:reasoning_results}

To validate the generality of our approach, besides the human-AI coordination benchmark Overcooked-AI, we further test our approach on some typical reasoning benchmarks to see whether utilizing multiple sessions can also yield benefits on tasks of other domains. In \ref{exp:reasoning}, we report the experimental results on \textit{Symbolic Manipulation} benchmark, and find that our approach does enhance the reasoning accuracy of LLMs and obtains superior performance compared to previous methods. The further results on \textit{Compositional Generalization} and \textit{Math Reasoning} are respectively presented in \ref{tab:multiple_llm_composition} and \ref{tab:multiple_llm_math}. 

On the \textit{Compositional Generalization} benchmark, we also reproduce the results of Chain-of-Thought and Least-to-Most using the latest GPT-3.5 model. We find that the replicated results differ from the results reported in the paper of Least-of-Most, for which we hypothesize the reason is the update of the GPT-3.5 model. However, when equally employing GPT-3.5, our method of utilizing $2$ sessions can be able to obtain accuracy results significantly higher than the baseline algorithms. Moreover, when we add the number of LLMs to $3$, our method exhibits even better reasoning performance, achieving accuracy of 90\%.

\begin{table}[t!]
    \centering
    \caption{Reasoning accuracy on the \textit{Compositional Generalization} benchmark.}
    \label{tab:multiple_llm_composition}
    \resizebox{\linewidth}{!}{
    \begin{tabular}{cccccc}
    \toprule
        \multirow{2}{*}{LLM-Model} & \multicolumn{5}{c}{Method} \\ \cmidrule{2-6}
        & Standard prompting & Chain-of-Thought & Least-to-Most & Ours (2 session) & Ours (3 session) \\ \midrule
        code-davinci-002  (original paper) & 16.7 & 16.2 & \textbf{99.7} & / & / \\ \midrule
        GPT-3.5 & 0 & 35 & 25 & 65 & \textbf{90} \\ \bottomrule
    \end{tabular}}
\end{table}

On the \textit{Math Reasoning} benchmark, we find that Standard prompting with the latest GPT-3.5 model achieves significantly higher accuracy than reported in the paper of Least-to-Most, which indicates that now the updated GPT-3.5 model can have been fine-tuned on similar datasets. For this reason, all methods do not show a very significant improvement to Standard prompting on this benchmark. Chain-of-Thought obtains the highest score on Non-Football (DROP), while our method performs slightly better on Football (DROP) and GSM8K.\newline

\begin{table}[htbp]
    \centering
    \caption{Reasoning accuracy on the \textit{Math Reasoning} benchmarks.}
    \label{tab:multiple_llm_math}
    \resizebox{\linewidth}{!}{
    \begin{tabular}{cccc}
    \toprule
    \multirow{2}{*}{Method (LLM-Model)} & \multicolumn{3}{c}{Benchmarks} \\ \cmidrule{2-4}
                             & Non-Football (DROP) & Football (DROP) & GSM8K \\ \midrule
        Standard prompting (original paper) & 58.78 & 62.73 & 17.06 \\ \midrule
        Chain-of-Thought   (original paper) & 74.77 & 59.59 & 60.87 \\ \midrule
        Least-to-Most   (original paper) & 82.45 & 73.42 & 62.39 \\ \midrule
        Standard prompting (GPT-3.5) & 85.00 & \textbf{80.00} & 70.00 \\ \midrule
        Chain-of-Thought  (GPT-3.5) & \textbf{90.00} & 65.00 & 70.00 \\ \midrule
        Least-to-Most  (GPT-3.5) & 85.00 & 70.00 & 65.00 \\ \midrule
        Ours  (GPT-3.5) & 85.00 & \textbf{80.00} & \textbf{75.00} \\ \bottomrule
    \end{tabular}}
\end{table}

\subsection{Ethical Statement}

In terms of the real human experiments, we have adopted effective practice to mitigate potential risks and ethical issues. Actually, the main risks for the human volunteers in our experiments are 1) personal information leakage and 2) time cost. For the former, we only invite the volunteers to participate in the test experiments without requirement for any extra personal information. Also, we will maintain strict confidentiality of the volunteers' identities. For the latter, we improve the interactive interface using the \texttt{Flask} framework to enhance the user experience and testing efficiency. Besides, we provide certain material compensation to the volunteers. 

For the boarder impact of our approach, as our work has not reached the application stage, it does not have a boarder impact currently. In the future, we will carefully consider the societal impact our method may bring.

\subsection{Prompt Engineering}\label{sec:prompt}

\textbf{Prompts for the Overcooked-AI.}

\begin{mdframed}[backgroundcolor=bg-gray]
\commenttext{==== Prompt for Session 1, to extract key information related to task planning tasks ====} \\
In a collaborative cooking game, you are an AI who needs to play the role of a chef with one human player. Before the game starts, humans will communicate with you, giving you human instructions. Please extract key information from human instructions, including 'Cooking Objectives' and' My Work '.\\
Making any dish requires using a pot and completing two tasks: 'Fetching vegetables' and 'Delivering food'. Therefore, if humans indicate that you need to make a certain type of food yourself, then you need both 'Fetching vegetables' and 'Delivering food'. Among them, 'Fetching vegetables' refers to placing an uncooked ingredient in a pot for the next step of work. 'Delivering food' refers to the delivery of food from a pot to the delivery port after it has been cooked.\\
If the cooking objective is tomato soup, then the ingredient to be placed in the pot is tomato; Similarly, if the cooking goal is onion soup, the ingredient to be prepared in the pot is onion.\\
\\
For Example 1: \\
The instructions for humans are: Please make tomato soup.\\
Your answer:\\
Cooking objectives: tomato soup\\
AI’s jobs: \\
Fetching vegetables: All pots.\\
Delivering food: All pots.\\
\\
For Example 2: \\
The instructions for humans are: Please make tomato soup, and you are only responsible for preparing tomatoes. Please take the tomatoes from the tomato spot on the right.\\
Your answer:\\
Cooking objectives: tomato soup\\
AI’s jobs: \\
Fetching vegetables: All pots.\\
Delivering food: Not mentioned.\\
\\
For Example 3: \\
The instructions for humans are: Please use the pot on the right to make onion soup. \\
Your answer:\\
Cooking objectives: onion soup\\
AI’s jobs: \\
Fetching vegetables: the pot on the right.\\
Delivering food: the pot on the right.\\
\\
For Example 4: \\
The instructions for humans are: Please use the pot on the left to make onion soup and be responsible for the delivery of the middle pot.\\
Your answer:\\
Cooking objectives: onion soup\\
AI’s jobs: \\
Fetching vegetables: the pot on the left.\\
Delivering food: the pot on the left + the middle pot.\\
\\
Now, the instructions for humans are: Please join me in making onion soup. You are only responsible for putting the onion into the pot and do not take onions from the onion dots below.
Please provide your answer by giving examples.
\end{mdframed}
\begin{mdframed}[backgroundcolor=bg-gray]
\commenttext{==== Prompt for Session 2, to clarify the rough work content ====} \\
In a collaborative cooking game, you need to play the role of a chef with one human player. To collaborate better with humans in the game, you need to plan your rough work content with humans before the game starts.\\
Firstly, you need to clarify the location of the pot mentioned in the key information. Then, based on the job description of AI in the key information, obtain the rough work content of AI. Finally, obtain the rough work content that humans need to complete.\\
\\
For Example 1:\\
The pot in the scene: (1,2), (1,3), (1,4)\\
Key information in human instructions: \\
Cooking objectives: onion soup\\
AI’s jobs: \\
(1) Fetching vegetables: the pot on the left.\\
(2) Delivering food: the pot on the left + the middle pot.\\
\\
Your answer:\\
the pot on the left is pot (1,2)\\
the middle pot is pot (1,3)\\
So, the rough work contents that AI need to do are:\\
(1) Fetch onions for pot at (1,2)\\
(2) Deliver onion soup for pot (1,2)\\
(3) Deliver onion soup for pot (1,3)\\
Correspondingly, the rough tasks that humans need to complete are:\\
(1) Fetch onions for pot at (1,3)\\
(2) Fetch onions for pot at (1,4)\\
(3) Deliver onion soup for pot (1,4)\\
\\
Now, the pot in the scene: (2,3), (1,3), (1,4)\\
Key information in human instructions: \\
Cooking objectives: tomato soup\\
AI’s jobs: \\
(1) Fetching vegetables: the pot below.\\
(2) Delivering food: other pots.\\
Please provide your answer by giving examples.\\
\end{mdframed}
\begin{mdframed}[backgroundcolor=bg-gray]
\commenttext{==== Prompt for Session 3, to refine the rough work content ====} \\
Please refine the rough work content based on the scenario information.\\
The rough work content is divided into two categories:\\
One is to pick up ingredients, which need to be refined to where to take the ingredients from and into which pot. \\
The second is to deliver food, which needs to be refined to the location from which the plate is taken, then the food is placed in the plate in which pot, and then the food loaded on the plate is sent to which delivery port.\\
In addition, you also need to consider whether there are restrictions on the items you can use in human instructions.\\
\\
For Example 1:\\
Human instructions: Please prepare onions.\\
The rough work content is: Pick up onions for pot (1,2)\\
Scenario information is:\\
Location of Tomatoes: (2,5), (3,5)\\
Location of Onions: (2,1), (3,1)\\
Location of the dining plate: (4,1), (4,5)\\
Location of the delivery port: (5,2)\\
Your answer:\\
There are no additional restrictions in the human instructions on where to take onions. \\
For the first onion position (2,1), its distance from the pot (1,2) is $| 2-1 |+| 1-2 |=1+1=2$. \\
For the second onion position (3,1), its distance from the pot (1,2) is $| 3-1 |+| 1-2 |=2+1=3$. \\
Therefore, I should choose a location (2,1) closer to the pot (1,2) to take the onion.\\
So, the refined work content is: Take the onion from position (2,1) and place it in the pot (1,2).\\
\\
For Example 2:\\
Human instructions: Please use the pot on the right to make tomato soup.\\
The rough work content is: Deliver tomato soup for pot to (1,3)\\
Scenario information is:\\
Location of Tomatoes: (2,5), (3,5)\\
Location of Onions: (2,1), (3,1)\\
Location of the dining plates: (4,1), (4,5)\\
Location of the delivery ports: (5,2)\\
Your answer:\\
There are no additional restrictions in the human instructions on where to pick up the plate and which delivery point to deliver it to. \\
For the first dining plate position (4,1), its distance from the pot (1,3) is $|4-1 |+| 1-3 |=3+2=5$. \\
For the second dining plate position (4,5), its distance from the pot (1,3) is $| 4-1 |+| 5-3 |=3+2=5$. \\
Therefore, I should choose a location (4,1) closer to the pot (1,3) to take a plate.
For the first delivery port (5,2), its distance from the pot (1,3) is $| 5-1 |+| 2-3 |=4+1=5$.\\
Therefore, I should choose the delivery port (5,2) closer to the pot (1,3) to deliver the food.\\
So, the refined work content is: Take the plate from (4, 1), then take the food from the pot (1, 3), and finally deliver it to the delivery port (5, 2).\\
\\
For Example 3:\\
Human instructions: Please prepare onions. You can only take onions from the onion dots below.\\
The rough work content is: Pick up onions for pot (1,2)\\
Scenario information is:\\
Location of Tomatoes: (2,5), (3,5)\\
Location of Onions: (2,1), (3,1)\\
Location of the dining plate: (4,1), (4,5)\\
Location of the delivery port: (5,2)\\
Your answer:\\
The human instructions require me to pick onions from the onion dots below.\\
the onion dots below is (3,1).\\
So, the refined work content is: Take the onion from position (3,1) and place it in the pot (1,2).\\
\\
Now, the rough work content is: Deliver tomato soup for pot (1,4)\\
Scenario information is:\\
Location of Tomatoes: (2,1), (2,5)\\
Location of Onions: (3,1), (3,5)\\
Location of the dining plates: (4,1), (4,5)\\
Location of the delivery ports: (5,2)\\
Please provide your answer by giving examples.
\end{mdframed}
\begin{mdframed}[backgroundcolor=bg-gray]
\commenttext{==== Prompt for Session 4, to calculate the approximate time required to execute each detailed work content ====} \\
There is currently a task in a grid world, please estimate the approximate time required to perform this task.\\
Among them, each move of a character requires one time step, and interacting with objects in the scene requires one time step.\\
For two types of work:\\
(1) Fetching vegetables: the approximate time is six times the time it takes to move the vegetables from their position to the pot position.\\
(2) Delivering food: the approximate time is from the position of the plate to the position of the pot, to the position of the delivery port, and then to the position of the plate.\\
\\
For Example 1:\\
The rough work content is: Pick up onions for pot (1,2)\\
The refined work content is: Take the onion from position (2,1) and place it in the pot (1,2).\\
Your answer:\\
Moving onions from (2,1) to (1,2) requires $|2-1|+|1-2|=1+1=2$ steps. \\
So, the approximate time is: $2 \times 6 = 12$ steps.\\
\\
For Example 2:\\
The rough work content is: Deliver tomato soup for pot (1,3)\\
The refined work content is: Take the plate from (4, 1), then take the food from the pot (1, 3), and finally deliver it to the delivery port (5, 2).\\
Your answer:\\
Moving from (4,1) to (1,3) requires $|4-1|+|1-3|=3+2=5$ steps. \\
Moving from (1,3) to (5,2) requires $|1-5|+|3-2|=4+1=5$ steps. \\
Moving from (5,2) to (4,1) requires $|5-4|+|2-1|=1+1=2$ steps. \\
So, the approximate time is: $5 + 5 +2 = 12$ steps.\\
\\
Now, the rough work content is: Deliver tomato soup for pot (1,4)\\
The refined work content is: Take the plate from (4, 5), then take the food from the pot (1, 4), and finally deliver it to the delivery port (5, 2).\\
Please provide your answer by giving examples.\\
\end{mdframed}
\begin{mdframed}[backgroundcolor=bg-gray]
\commenttext{==== Prompt for Session 5, to adjust the order of work tasks ====} \\
Please adjust the order of execution for the rough work content that needs to be completed.\\
Please note that there are two rough job descriptions: picking up vegetables and delivering food. However, for the same pot, delivery work can only be carried out 20 time steps after completing the vegetable picking work.\\
\\
For Example 1:\\
The rough work contents are: \\
(1) Fetch onions for pot at (1,2), 12 steps\\
(2) Deliver onion soup for pot (1,2), 10 steps\\
(3) Fetch onions for pot at (1,3), 18 steps\\
Your answer:\\
Due to the fact that the execution of the work 'Delivery on soup for pot (1,2)' requires 20 time steps after completing the 'Pick up onions for pot (1,2)', in order to fully utilize the waiting time, other work should be performed during this period.\\
Therefore, the work sequence should be adjusted to:\\
(1) Fetch onions for pot at (1,2), 12 steps\\
(2) Fetch onions for pot at (1,3), 18 steps\\
(3) Deliver onion soup for pot (1,2), 10 steps\\
\\
Now, the rough work contents are:\\
(1) Fetch onions for pot at (1,2), 12 steps\\
(2) Deliver onion soup for pot (1,2), 10 steps\\
(3) Fetch onions for pot at (1,3), 8 steps\\
(4) Fetch onions for pot at (1,4), 10 steps\\
Please give me your answer as the example.
\end{mdframed}

\textbf{We also provide an example of re-plan due to human feedback below}.

\begin{mdframed}[backgroundcolor=bg-gray]
\commenttext{==== First round ====} \\
Human: Please join me in making onion soup.\\
AI: \\
\commenttext{[Output of LLM1]}\\
Cooking objectives: onion soup\\
AI’s jobs:\\
Fetching vegetables: All pots.\\
Delivering food: All pots.\\
\\
\commenttext{[Output of LLM2]}\\
The pot in the scene is pot (3,6).\\
Based on the job description of AI in the key information, the rough work content for AI would be:\\
(1) Fetch onions for pot at (3,6)\\
(2) Deliver onion soup for pot (3,6)\\
Since there is only one pot in the scene, the rough tasks that the human player needs to complete would be: None.\\
...\\
\commenttext{[Ouptut of LLM5]}\\
The work content and execution sequence of AI:\\
(1) Fetch onions for pot (3,6): Take the onion from position (5,5) and place it in the pot (3,6).\\
(2) Deliver onion soup for pot (3,6): Take the plate from (1, 4), then take the food from the pot (3, 6), and finally deliver it to the delivery port (3, 1).\\
The work content and execution sequence of Human:\\
None\\
\\
\textcolor{dark-yellow}{[Human feedback]}\\
\llmcompletion{You are only responsible for putting the onion into the pot,}\\
\llmcompletion{and do not take onions from the onion dots below.}\\
\\
\commenttext{==== Second round ====} \\
...\\
AI:\\
\commenttext{[Ouput of Session 1]}\\
Cooking objectives: onion soup\\
AI’s jobs:\\
Picking up vegetables: All pots.\\
Delivering food: Not mentioned.\\
...\\
\commenttext{[Ouput of Session 5]}\\
The work content and execution sequence of AI:\\
(1) Fetch onions for pot at (3,6): Take the onion from position (1,7) and place it in the pot (3,6).\\
The work content and execution sequence of Human:\\
(1) Deliver onion soup for pot (3,6): Take the plate from (1, 4), then take the food from the pot (3, 6), and finally deliver it to the delivery port (3, 1).
\end{mdframed}

\end{document}